\pdfoutput=1 

\documentclass[10pt,twocolumn,letterpaper]{article}

\usepackage{iccv}
\usepackage{times}
\usepackage{epsfig}
\usepackage{graphicx}
\usepackage{amsmath}
\usepackage{amssymb}

\usepackage{caption}
\usepackage{subcaption}

\usepackage{algorithm}
\usepackage{algorithmic}

\makeatletter
\@namedef{ver@everyshi.sty}{}
\makeatother

\usepackage{tikz}
\usepackage{makecell}
\usepackage{multirow}
\usepackage{gensymb}

\usepackage[accsupp]{axessibility}

\newcommand\norm[1]{\lVert#1\rVert}

\newcommand{\tableFont}{\fontsize{8pt}{9.5pt} \selectfont}

\usepackage[breaklinks=true,bookmarks=false]{hyperref}
\iccvfinalcopy 

\def\iccvPaperID{6272} 

\ificcvfinal\fi

\begin{document}

\title{Surface Normal Clustering for Implicit Representation of Manhattan Scenes}

\author{Nikola~Popovic$^{1}$, Danda~Pani~Paudel$^{1,2}$, Luc~Van~Gool$^{1,2}$\\
$^{1}$Computer Vision Laboratory, ETH Zurich, Switzerland\\
$^{2}$INSAIT, Sofia University, Bulgaria\\
{\tt\small \{nipopovic, paudel, vangool\}@vision.ee.ethz.ch}
}
\date{}

\maketitle
\ificcvfinal\thispagestyle{empty}\fi

\begin{abstract}
Novel view synthesis and 3D modeling using implicit neural field representation are shown to be very effective for calibrated multi-view cameras. Such representations are known to benefit from additional geometric and semantic supervision. Most existing methods that exploit additional supervision require dense pixel-wise labels or localized scene priors. These methods cannot benefit from high-level vague scene priors provided in terms of scenes' descriptions. In this work, we aim to leverage the geometric prior of Manhattan scenes to improve the implicit neural radiance field representations. More precisely, we assume that only the knowledge of the indoor scene (under investigation) being Manhattan is known -- with no additional information whatsoever -- with an unknown Manhattan coordinate frame. Such high-level prior is used to self-supervise the surface normals derived explicitly in the implicit neural fields. Our modeling allows us to cluster the derived normals and exploit their orthogonality constraints for self-supervision. Our exhaustive experiments on datasets of diverse indoor scenes demonstrate the significant benefit of the proposed method over the established baselines. The source code is available at ~\url{https://github.com/nikola3794/normal-clustering-nerf}.
\end{abstract}
\section{Introduction}

\begin{figure*}[t]
\centering
\def\svgwidth{\textwidth}
 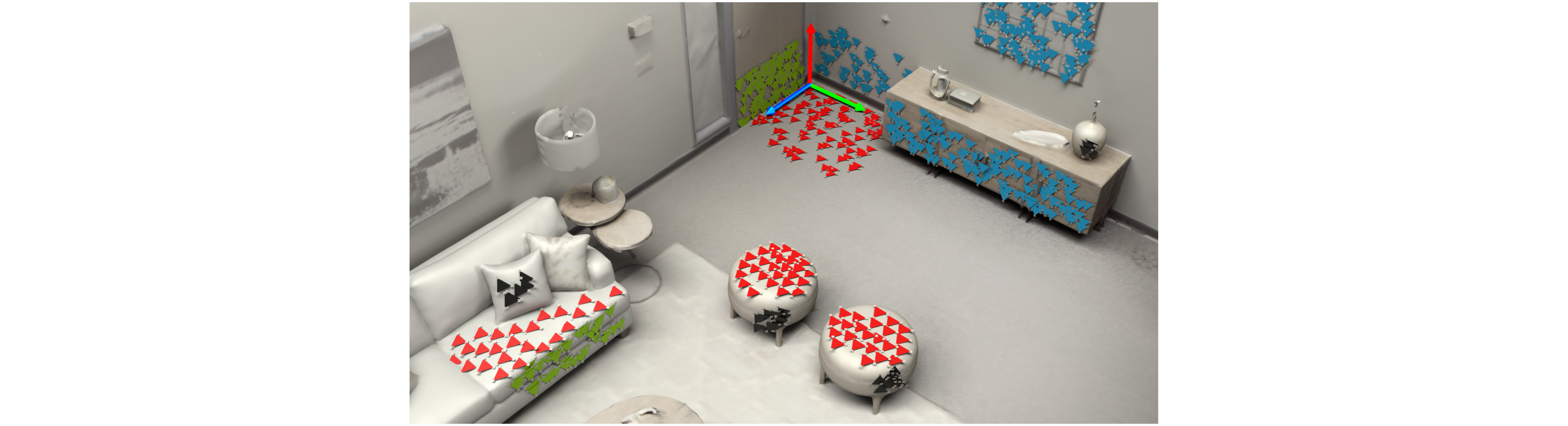
 \vspace{-2mm}
\caption{\textbf{Illustration of the proposed method.} We compute one surface normal for each ray triplet, using 3D surface points derived from rendered depths (left). The computed normals (from many ray triplets) are clustered to obtain the MF (middle). The non-Manhattan and noisy surfaces are handled by a robust orthogonal normals search, which is later used for self-supervision through the Manhattan prior (right). }
\vspace{-3mm}
\label{fig:teaser}
\end{figure*}

\label{sec:intro}
Above 80\% images ever taken are estimated to involve human-made architectural structures, with a substantial share of indoor scenes~\cite{fernandez2020indoor}. These scenes often exhibit strong structural regularities, including flat and texture-poor surfaces in some axis-aligned Cartesian coordinate system -- also known as Manhattan world~\cite{coughlan1999manhattan,furukawa2009manhattan}.  Paradoxically, these regularities may hinder the 3D modeling process if it is unaware of the human-made scene priors. In fact, several computer vision works have even benefited from the knowledge of the Manhattan world for the task of 3D scene reconstruction~\cite{vanegas2010building,furukawa2009manhattan},  camera localization~\cite{li2018monocular}, self-calibration~\cite{wildenauer2012robust}, and more~\cite{straub2017manhattan, singh2010visual, purkait2017rolling, zhai2016detecting}.

For calibrated multi-view cameras, 3D inversion using implicit neural representations~\cite{mildenhall2020nerf, wang2021neus, yariv2021volume} is becoming increasingly popular due to their remarkable performance and recent efficiency developments~\cite{yu2021plenoctrees,reiser2021kilonerf,mueller2022instant,xie2022neural}. Meanwhile, such representations are known to benefit from additional supervision in the form of depth~\cite{azinovic2022neural,deng2022depth,roessle2022dense}, normals~\cite{li2022nerf}, semantics~\cite{vora2021nesf,zhi2021place, kundu2022panoptic,jain2021putting}, local-regularization~\cite{wang2022neuris,verbin2022ref,niemeyer2022regnerf}, local planar patches~\cite{lin2022neurmips}, or their combinations~\cite{yu2022monosdf,guo2022neural,lin2022neurmips}. In this context, a notable recent work ManhattanSDF~\cite{guo2022neural} demonstrates the benefit of exploiting the high-level geometric prior for structured scenes. More precisely,  ManhattanSDF~\cite{guo2022neural} uses the known semantic regions to impose the planar geometry prior of floors and walls under the Manhattan scene assumption. During this process, the exact normal of the floor and partial normals of the walls are assumed to be known, with respect to the camera coordinates. 

We aim to improve the 3D neural radiance field representations for calibrated multi-view cameras in indoor Manhattan scenes, with no further assumptions. In other words, we consider that the structural and semantic information is not available, unlike  ManhattanSDF~\cite{guo2022neural}. In addition to the floor and walls used in ManhattanSDF~\cite{guo2022neural}, we wish to exploit the Manhattan prior of many other indoor scene parts (\eg tables and wardrobes). More importantly, we consider that the Manhattan coordinate frame is also unknown. Our assumptions (of unknown semantics and Manhattan frame) on one hand make our setting very practical. On the other hand, those assumptions make the problem of exploiting the Manhattan scene prior for 3D inversion very challenging. 

The virtue of the Manhattan world assumption comes from its simplicity, allowing us to intuitively reason about the geometry of a wide range of complex scenes/objects such as cities, buildings, and furniture. 
However, such reasoning often requires the axis-aligned Cartesian coordinate frame, also known as the Manhattan frame (MF), to be known~\cite{straub2017manhattan}. 
Unfortunately, recovering the Manhattan frame directly from images is not an easy task~\cite{coughlan1999manhattan, straub2017manhattan}. 
Therefore, several methods have been developed until  recently~\cite{denis2008efficient,bazin2012globally,straub2015real,joo2016globally,ge2021globally} to recover the Manhattan frame, relying upon the known 3D reconstruction or image primitives (eg. lines, planes). 
We wish to exploit the Manhattan prior for improving the 3D representation, without needing to know MF beforehand. Instead, our experiments reveal that knowing the MF beforehand offers no additional benefit in  Manhattan-prior aware radiance field representation. 


In this work, we propose a method that jointly learns the Manhattan frame and neural radiance field, from calibrated multi-view in indoor Manhattan scenes, in an end-to-end manner using the recent efficient backbone of  InstantNGP~\cite{mueller2022instant}. The proposed method requires no additional information to exploit the Manhattan prior and relies on the explicitly derived normals in the implicit neural fields. We use batches of three neighboring rays, whose effective surface's local piece-wise planarity is assumed to derive the explicit normals by algebraic means.  In pure, complete, and enclosed Manhattan scenes, these normals form six clusters corresponding to three orthogonal and other three parallel counterpart planes. However real scenes consist of non-Manhattan scene parts and missing planes. Therefore, we use a robust method that uses minimal three orthogonal clusters to recover the sought Manhattan frame. As in the literature, we seek a rotation matrix whose entries are directly derived from the orthogonal clusters of normals, to align the Manhattan frame. The recovered MF is then used to encourage the derived normals to be axis-aligned for self-supervision. A graphical illustration of our method is presented in Figure~\ref{fig:teaser}. Our extensive experiments demonstrate the robustness and utility of our method in improving the implicit 3D in neural radiance field representations.        

The major contributions of our paper are listed below:

\noindent  -- We address the problem of exploiting the structured-scene knowledge without requiring any dense or localized scene priors, for the first time in this paper.   \\
\noindent  -- We present a method that successfully exploits the Manhattan scene prior with an unknown Manhattan frame. The proposed method also recovers the unknown frame. \\
\noindent  -- We demonstrate the robustness and utility of our method on three indoor datasets, where our method improves the established baselines significantly. These datasets consist of 200+ scenes, making our method tested in significantly more scenes than the state-of-the-art methods.

\section{Related Works}
\label{sec:relatedWorks}
\noindent\textbf{Implicit neural representation of 3D:} 
Since the foundational work of \emph{Mildenhall et al.}~\cite{mildenhall2020nerf},  the implicit neural representation of 3D scenes has advanced on various fronts. These fronts include representation~\cite{yariv2021volume,suhail2022light}, generalization~\cite{wang2021nerf,park2021nerfies,lin2021barf,martin2021nerf, sajjadi2022scene}, generation~\cite{niemeyer2021giraffe,chan2022efficient}, and efficient methods~\cite{reiser2021kilonerf,yu2021plenoctrees,mueller2022instant}. We rely on a recent method InstantNGP~\cite{mueller2022instant}  developed by \emph{Mueller et al.}, as our backbone. Our choice is made primarily based on the computational efficiency during both training and inference. Thanks to the offered computational efficiency, we are able to conduct large-scale experiments on several scenes.

\noindent\textbf{Auxiliary supervision methods:} In addition to the images, other inputs such as depth~\cite{rosinol2022nerf,sucar2021imap, deng2022depth}, semantics~\cite{zhi2021place,jain2021putting,kundu2022panoptic,sun2022fenerf}, normal~\cite{yang2022ps,yang2022s}, and their combinations~\cite{yu2022monosdf,guo2022neural,lin2022neurmips} are shown to be beneficial on improving the neural radiance field representation. In this regard, these auxiliary supervisions often use ground-truth labels. It is needless to say that the need for ground-truth supervision is not desired, whenever possible. Therefore, recent methods use labels predicted by some pre-trained networks~\cite{yu2022monosdf} or recovered from the structure-from-motion (SfM) pipeline~\cite{deng2022depth,lin2022neurmips}.
One notable work ManhattanSDF~\cite{guo2022neural} exploits the Manhattan prior without requiring any SfM reconstruction. However, ManhattanSDF requires (a) semantics of the scene parts and (b)  the Manhattan frame, to be known.
We argue that such labels required for auxiliary supervision cannot always be relied upon, due to domain gaps, poor reconstruction of texture-less regions, and additional computational needs, to list a few. Therefore, we do not use any additional labels for auxiliary supervision.


\noindent\textbf{Manhattan frame estimation:} Since the early works of \emph{Bernard}~\cite{barnard1983interpreting}, Manhattan structure reasoning is done directly on/from images by detecting the so-called the vanishing-points (VPs)~\cite{mclean1995vanishing,van19983d}. In fact, the problem of detecting three orthogonal VPs is equivalent to finding the MF in 3D for the calibrated multi-view setting~\cite{bazin2012branch}. Note that the knowledge of Manhattan structure has been used in several computer vision works~\cite{vanegas2010building,furukawa2009manhattan, li2018monocular,wildenauer2012robust,straub2017manhattan, singh2010visual, purkait2017rolling, zhai2016detecting}. When unknown, most methods implicitly or explicitly estimate the MF to leverage the Manhattan scene prior. In~\cite{straub2017manhattan}, \emph{Straub et al.} have demonstrated that the MF can be efficiently represented in and recovered from the space of surface normals. We use a similar formulation as~\cite{straub2017manhattan}, using the surface normals derived explicitly from the implicit neural fields.   


\section{Background and Notations}
\label{sec:background}
Manhattan frame (MF) is a coordinate system that is defined by the structure building orthogonal planes of Manhattan scenes. We consider unknown MF since the scene planes and their geometric relationships are unknown. Let the unknown MF differs from the world-frame (WF), used for the 3D representation, by rotation $\mathsf{R}\in SO(3)$. We denote three orthogonal axes in MF by $\mathcal{E}=\{\mathsf{e}_x,\mathsf{e}_y,\mathsf{e}_z\}$.  Without loss of generality, let the axes' coordinates be $\mathsf{e}_x= [1,0,0]^\intercal$, $\mathsf{e}_y= [0,1,0]^\intercal$, and $\mathsf{e}_z= [0,0,1]^\intercal$. Note that these axes align with the normals of the respective scene building planes, in the WF. Let $\mathcal{N}=\{\mathsf{n}_i\}_{i=1}^m$ be a set of 3D normals of all the scene planes. Then, the rotation $\mathsf{R}$, from world to Manhattan frame, aligns the normals  $\mathsf{n}\in\mathcal{N}$ to the Manhattan axes  $\mathsf{e}\in\mathcal{E}$, i.e,  $\mathsf{Rn}_i\in\mathcal{E},\,\forall \mathsf{n}_i\in\mathcal{N}$.

We are interested to recover $\mathsf{R}$ from a set of normals $\mathcal{N}$. 
For this, the above set-to-set assignment alone is not sufficient. 
This requires the element-wise assignment between sets $\mathcal{N}$ and $\mathcal{E}$.
To do so, we divide the set $\mathcal{N}$ into three orthogonal subsets $\mathcal{N}_x$, $\mathcal{N}_y$, and $\mathcal{N}_z$. Now, for any triplet of $\{\mathsf{n}_x,\mathsf{n}_y,\mathsf{n}_z\}$ from the corresponding orthogonal subsets, we aim to establish the condition ${[\mathsf{e}_x,\mathsf{e}_y,\mathsf{e}_z] = \mathsf{R}[\mathsf{n}_x,\mathsf{n}_y,\mathsf{n}_z]}$. Note that the assignment condition results into $\mathsf{R} = [\mathsf{n}_x,\mathsf{n}_y,\mathsf{n}_z]^\intercal$. Therefore, the problem of recovering MF from a given set of normals boils down to finding three normals from orthogonal subsets. At this point, one issue regarding robustness remains pending. More precisely, we wish to recover $\mathsf{R}$ for a noisy set of normals $\mathcal{N}$, with potentially overwhelmingly many outliers.

We use a robust method to recover $\mathsf{R}$  from given normals $\mathcal{N}$. In principle, $\mathsf{R}$ can be recovered from minimal two orthogonal normals, with one additional normal for disambiguation by sign correction. The recovered rotation can then be validated by consensus for robust recovery~\cite{straub2017manhattan}. Alternatively,  one can also perform the robust estimation of the orthogonal subsets of $\mathcal{N}$, followed by solving  ${[\mathsf{e}_x,\mathsf{e}_y,\mathsf{e}_z] = \mathsf{R}[\mathsf{n}_x,\mathsf{n}_y,\mathsf{n}_z], \forall \mathsf{n}_x\in\mathcal{N}_x,\mathsf{n}_y\in\mathcal{N}_y,\mathsf{n}_z\in\mathcal{N}_z},$ for $\mathsf{R}\in SO(3)$. For computational reasons, we estimate the robust subsets by clustering. The orthogonal subsets are then obtained by choosing the three most orthogonal clusters. Later, the obtained cluster centers are used to estimate the MF (or to enforce its existence) represented by $\mathsf{R}$. Please refer to Section~\ref{subSec:robustNormals} for more details.



\section{Method}
\label{sec:method}
\begin{figure*}[t]
\centering
\def\svgwidth{0.95\textwidth}
 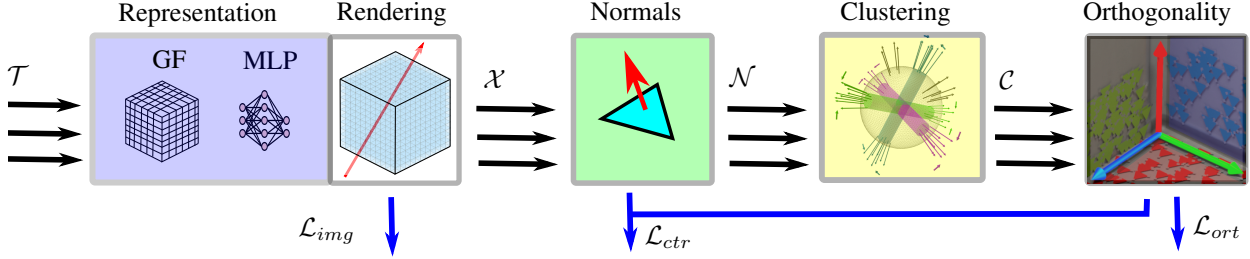
 \vspace{-2mm}
\caption{ \textbf{The complete pipeline of our method.} We use grid features (GF) and an MLP to represent the radiance field. The explicit normals are derived using the depths obtained from volume rendering. The Manhattan scene prior is exploited by clustering the estimated normals to enforce their orthogonality.}
\vspace{-3mm}
\label{fig:flowDiagram}
\end{figure*}

From a set of calibrated images, we model the 3D scene using the neural radiance field representation. In the process, we wish to exploit the Manhattan scene prior, without the knowledge of the Manhattan frame. This problem is addressed by jointly optimizing for the neural radiance field and the Manhattan frame estimation. The complete pipeline of our method is illustrated in Figure~\ref{fig:flowDiagram}. As shown, our method consists of three units: (a) Explicit normal modeling, (b) Robust estimation of orthogonal normals, and (3) Self-supervision by Manhattan prior. In the first stage, we shoot a batch of three rays which allows us to estimate the surface normal using an algebraic method. As the estimated normals are bound to be noisy (with outliers), we perform their clustering to obtain the most orthogonal clusters (representing the Manhattan frame) in a robust manner. The obtained clusters are then used to estimate the sought MF, in the form of a rotation matrix, using the method discussed in Section~\ref{sec:background}. In the final step, we use the estimated Manhattan frame to encourage the close-by normals to be Manhattan-like and to enforce a stricter orthogonality constraint for self-supervision.
In the following, we will present the details of three units of our method in different Subsections.

\subsection{Explicit Normal Modelling}
At any given view, we consider the color $c= C_\theta(\mathsf{r})$ and depth $d= D_\theta(\mathsf{r})$ for a ray $\mathsf{r}$ emanating from the corresponding camera center $\mathsf{o}$ is obtained using the volumetric rendering of the implicit neural radiance field, represented by a neural network parameterized by $\theta$ as in~\cite{mildenhall2020nerf}. Then, the location of the 3D surface point is given by,  $ \mathsf{x} = \mathsf{o}+d.\mathsf{r}.$
We process triplets of three neighboring rays. Let $\mathcal{T} = \{\mathsf{r}_1,\mathsf{r}_2,\mathsf{r}_3\}$ be such a triplet, whose corresponding surface points are given by $\mathcal{X} = \{\mathsf{x}_1,\mathsf{x}_2,\mathsf{x}_3\}$.  Now, for the ray triplet $\mathcal{T}$, we obtain the explicit surface normal $\mathsf{n}$ with the help of the point triplet $\mathcal{X}$, using the following mapping and algebraic operation,
\begin{equation}
    \mathcal{T}\rightarrow \mathcal{X} \rightarrow \mathsf{v} = \scalebox{0.75}[1]{$(\mathsf{x}_1-\mathsf{x}_2)\times (\mathsf{x}_1-\mathsf{x}_3)$}\rightarrow \mathsf{n} =  \frac{\text{sign}(\mathsf{o}^\intercal \mathsf{v})\mathsf{v}}{\norm{\mathsf{v}}}.\label{eq:mappingAndNormal}
\end{equation}
Note that $\mathsf{v}$ is a vector orthogonal to the plane passing through 3D points in triplet $\mathcal{T}$. We obtain the oriented normal $\mathsf{n}$ by normalizing  $\mathsf{v}$ and correcting its sign by ensuring that the camera center $\mathsf{o}$ is in front of the estimated plane. A graphical illustration of the surface normal estimation is provided in Figure~\ref{fig:teaser} on the left. We select a random set of ray triplets $\{\mathcal{T}_i\}_{i=1}^m$ from multiple cameras. These ray triplets provide us the surface point triplets $\{\mathcal{X}_i\}_{i=1}^m$. We use  point triplets to explicitly derive the  surface normals $\mathcal{N}=\{\mathsf{n}_i\}_{i=1}^m$, using~\eqref{eq:mappingAndNormal}. These normals are later used to recover the unknown Manhattan frame. The simplicity of the explicit normals computed in this paper makes them easy to compute and handle. If needed, normals of different sizes of surface regions could also be computed similarly. 

\subsection{Robust Estimation of Orthogonal  Normals}
\label{subSec:robustNormals}
We are interested to recover the MF from a set of noisy surface normals $\mathcal{N}=\{\mathsf{n}_i\}_{i=1}^m$. Recall Section~\ref{sec:background}, the MF can be obtained by robustly recovering a set of three orthogonal normals from $\mathcal{N}$. To do so, we first cluster $\mathcal{N}$ into $k$ clusters $\{\mathcal{C}_i\}_{i=1}^k$, by using the the well-known $k$-means clustering algorithm~\cite{bishop2006PRML}. During the clustering, every centroid is $L_2$ normalized after each iteration, to ensure that they are unit vectors to represent surface normals. In a perfect Manhattan world, there exist only six clusters corresponding to the three orthogonal MF axes $\mathcal{E}=\{\mathsf{e}_x,\mathsf{e}_y,\mathsf{e}_z\}$ and their parallel counterparts. However, real scenes consist of surfaces of different orientations.
An additional source of non-Manhattan normals comes from the inaccuracy in the normal estimation. Therefore, we use a clustering technique to robustly recover the orthogonal normals.  In the following, we present how the orthogonal clusters are selected from the set of clusters  $\{\mathcal{C}_i\}_{i=1}^k$. Note from Section~\ref{sec:background}, the selected orthogonal clusters are considered to represent MF defining sets $\mathcal{N}_x$, $\mathcal{N}_y$, and $\mathcal{N}_z$. We proceed by first selecting three orthogonal clusters, say $\mathcal{N}_1$, $\mathcal{N}_2$, and $\mathcal{N}_3$. Later, we assign them to $\mathcal{N}_x$, $\mathcal{N}_y$, and $\mathcal{N}_z$ to recover MF.

For notational ease, we pair clusters and centroids as ${\mathcal{U}=\{(\mathcal{C}_i,\mathsf{c}_i)\}_{i=1}^k}$, where the centroids are computed by taking the average across the corresponding cluster such that  ${\mathsf{c}_i = \frac{1}{|\mathcal{C}_i|}\sum_{\mathsf{n} \in \mathcal{C}_i} \mathsf{n}}$, followed by normalizing to a unit vector $\mathsf{c}_i = \frac{\mathsf{c}_i}{\norm{\mathsf{c}_i}}$. 
 Leveraging the Manhattan prior through the assumption that Manhattan surfaces dominate the scene, we pick $\mathsf{n}_1 = \mathsf{c}_{f}$, where $f=\arg \max_{i} |\mathcal{C}_i|$. In other words, we pick the centroid of the largest cluster, as one of the MF axes. Then, we obtain $\mathsf{n}_2 = \mathsf{c}_{s}$ and $\mathsf{n}_3 = \mathsf{c}_{t}$ as a solution of the optimization problem ${s, t = \arg \min_{i, j} |\mathsf{c}_i^{\intercal} \mathsf{n}_1| + |\mathsf{n}_1^{\intercal} \mathsf{c}_j| + |\mathsf{c}_i^{\intercal} \mathsf{c}_j|}$. In other words, we find two additional cluster centroids providing the most orthogonal triplet. We further merge all selected clusters with their opposites. This is achieved by comparing all cluster pairs. Whenever two centroids are nearby, but opposite in sign, the corresponding clusters are merged with the appropriate sign correction. The procedure of finding $\{\mathcal{N}_1, \mathcal{N}_2, \mathcal{N}_3\}$ is illustrated in Figure~\ref{fig:teaser} (middle and right), and is summarized in Algorithm~\ref{al:findManhattanFrame} from step 1--4. Although the orthogonal clusters with their centroids are sufficient to exploit the desired Manhattan prior, we may wish to recover the MF in the form of a rotation matrix. As such, any arrangement of $\{\mathsf{n}_1,\mathsf{n}_2,\mathsf{n}_3\}$ as a valid rotation matrix offers us a valid MF. We may however often be interested to recover the one which is closest to the world frame. For this reason, we suggest keeping the largest cluster's centroid paired to the closes canonical axis in $\mathcal{E}$. Similarly, we also align one more axis, whereas the last remaining axis gets paired by default. We summarize how to recover MF closest to the world frame in Algorithm~\ref{al:findManhattanFrame} in steps 5--7.

\begin{figure}[t]
\vspace{-4mm}
\begin{algorithm}[H]
\small
\caption{\small $(\mathcal{N}_1, \mathcal{N}_2, \mathcal{N}_3, \mathsf{R})$ = findManhattanFrame($\mathcal{N}$)}
\label{al:findManhattanFrame}
\begin{algorithmic}
 \STATE 1. Cluster normals $\mathcal{N}$ into, $\mathcal{U} = \{(\mathcal{C}_i,\mathsf{c}_i)\}_{i=1}^k$ using k-means. 
 \STATE 2. For the largest cluster $\mathcal{C}\in\mathcal{U}$, assign $(\mathcal{C},\mathsf{c})\rightarrow (\mathcal{N}_1,\mathsf{n}_1)$.
 \STATE 3. Find $\mathcal{C}_s\in\mathcal{U},\mathcal{C}_t\in\mathcal{U}$ minimizing
 ${ |\mathsf{c}_s^{\intercal} \mathsf{n}_1| + |\mathsf{n}_1^{\intercal} \mathsf{c}_t| + |\mathsf{c}_s^{\intercal} \mathsf{c}_t|}$.
 \STATE 4. Assign $(\mathcal{C}_s,\mathsf{c}_s)\rightarrow (\mathcal{N}_2,\mathsf{n}_2), ( \mathcal{C}_t,\mathsf{c}_t)\rightarrow(\mathcal{N}_3,,\mathsf{n}_3)$.
 \STATE 5. \scalebox{0.95}[1]{$|\mathsf{e}_z^\intercal\mathsf{n}_1|\leq\frac{1}{\sqrt{2}}?\mathsf{n}_1\rightarrow \mathsf{n}_z :(|\mathsf{e}_y^\intercal\mathsf{n}_1|\leq\frac{1}{\sqrt{2}}?\mathsf{n}_1\rightarrow \mathsf{n}_y:\mathsf{n}_1\rightarrow \mathsf{n}_x).$}
 \STATE 6. Perform the remaining $\{\mathsf{n}_1, \mathsf{n}_2,\mathsf{n}_3\}\rightarrow \{\mathsf{n}_x,\mathsf{n}_y,\mathsf{n}_z\}$ \\ \hspace{3mm} as in step 5 to the closest corresponding canonical axes. 
 \STATE 7. Return $\mathcal{N}_1,\mathcal{N}_2,\mathcal{N}_3$, and $\mathsf{R}=[\mathsf{n}_x,\mathsf{n}_y,\mathsf{n}_z]^\intercal$.
\end{algorithmic}
\end{algorithm}
\vspace{-9mm}
\end{figure}

\subsection{Self-supervision by Manhattan Prior}
The supervision of the implicit neural radiance field, parameterized by $\theta$, is achieved through jointly optimizing three loss terms. The first one is the photometric loss computed as follows, $\mathcal{L}_{img} = \frac{1}{|\mathcal{R}|} \sum_{\mathsf{r} \in \mathcal{R}} \norm{C_\theta(\mathsf{r}) - C(\mathsf{r})}_{2}^{2}$,
where $C(\mathsf{r})$ is the ground-truth color, and $\mathcal{R}$ is the set of rays going through sampled pixel triplets. This loss term is responsible for facilitating the learning of the implicit 3D representation of the investigated scene~\cite{mildenhall2020nerf}. 

\noindent\textbf{Losses from Manhattan prior:}
We exploit the Manhattan scene prior for self-supervision, in order to improve the implicit 3D representation without additional ground truth labels. We do so by using the clusters $\{\mathcal{N}_1, \mathcal{N}_2, \mathcal{N}_3\}$ obtained from Algorithm~\ref{al:findManhattanFrame}. More precisely, for cluster-centroid pairs $\{(\mathcal{N}_i,\mathsf{n}_i)\}_{i=1}^3$  we use the following two losses,
\begin{align}
    \mathcal{L}_{ctr} &= \frac{1}{3}\sum_{i=1}^3
    \frac{1}{|\mathcal{N}_i|} 
    \sum_{\mathsf{n} \in \mathcal{N}_i}
    \norm{1-\mathsf{n}_i^{\intercal} \mathsf{n}}_{1}
    +
    \norm{\mathsf{n}_i - \mathsf{n}}_{1},\\
    \mathcal{L}_{ort} &= \frac{1}{3} (
    |\mathsf{n}_1^{\intercal} \mathsf{n}_2| + 
    |\mathsf{n}_1^{\intercal} \mathsf{n}_3| + 
    |\mathsf{n}_2^{\intercal} \mathsf{n}_3|).
\end{align}
Here, the loss $\mathcal{L}_{ctr}$ encourages tighter clusters, while the loss $\mathcal{L}_{ort}$ enforces the orthogonality among the three clusters. 
The final loss used to optimize $\theta$ is then given by
$\mathcal{L} = \mathcal{L}_{img} + \lambda_{ctr}\mathcal{L}_{ctr} + \lambda_{ort}\mathcal{L}_{ort}$,
where $\lambda_{ctr}$ and $\lambda_{ort}$ are the hyperparameters. Please, refer to Figure~\ref{fig:flowDiagram} for a schematic summary of all three losses.

\section{Experiments}
\label{sec:Experiments}

\subsection{Baselines, Metrics, and Implementation Details}
\noindent\textbf{InstantNGP~\cite{mueller2022instant} (baseline):} This method represents the scene as a multi-resolution voxel grid and leverages a hash table of trainable feature vectors, which are used to represent grid elements. The representation is further processed with a small MLP. This allows InstantNGP to be very computationally efficient, requiring in our case around 30 minutes per scene to train and evaluate. 
Therefore, we use it as the backbone of our method and as our baseline.

\noindent\textbf{ManhattanDF~\cite{guo2022neural}:} This method exploits the Manhattan prior by supervising the explicit normals of the floors to align with the known $\mathsf{n}_z$, as well as by supervising normals of the walls to align with two learned orthogonal axes which are also orthogonal to $\mathsf{n}_z$. To achieve this, apart from the RGB images, this method relies on knowing the wall and floor semantics, as well as knowing the exact floor axis $\mathsf{n}_z$ (MF partially known). We implemented it on top of the InstantNGP backbone with density field estimation.


\noindent\textbf{Ours:} We implement our proposed method from Section~\ref{sec:method} on top of InstantNGP with density field estimation.

\noindent\textbf{Ours (MF known):} We modify our method by assuming that the MF is fully known. We do so by adding additional loss terms $\mathcal{L}_{man_i} = \norm{1-\mathsf{n}_i^{\intercal} \mathsf{m}}_{1} + \norm{\mathsf{n}_i - \mathsf{m}}_{1}$ for each orthogonal cluster centroid $\mathsf{n}_i$ and its closest MF axis $\mathsf{m}$ (or a closer opposite counterpart). Thus, we explicitly guide the orthogonal triplet to align with the known MF.


\noindent\textbf{Metrics:} To quantitatively evaluate novel view rendering, we use peak signal-to-noise ratio (PSNR) and the structural similarity index (SSIM)~\cite{wang2004ssim}. To evaluate the extracted surface normals of novel rendered views, we use the median angular error. To partially evaluate the quality of the learned implicit 3D structure, we utilize the mean absolute error (MAE) and the root mean square error (RMSE) on the depth obtained by volume rendering. Finally, to evaluate the recovered MF, we calculate the absolute error between the yaw, pitch, and roll angles of the recovered frame and the MF. All metrics are averaged across scenes, except for the yaw, pitch, and roll errors for which the median is reported.

\noindent\textbf{Implementation Details:} 
We turn on $\mathcal{L}_{ort}$ and $\mathcal{L}_{centr}$ after $500$ steps and linearly increase their weights to the specified values over the next $2500$ steps. Also, we randomly sample rays for one-third of every batch size and select their left and upper neighbor to form a triplet to facilitate obtaining explicit surface normals. For other implementation details, please refer to the supplementary material.

Note that in addition to RGB images, ManhattanDF~\cite{guo2022neural} requires knowing the wall and floor semantics, and the exact floor axis $\mathsf{n}_z$.
In contrast, our method only requires RGB images and the assumed Manhattan prior to holding true. 
Therefore, this comparison aims to get a better insight into what can be achieved without leveraging additional labels.

\subsection{Datasets}

\begin{table*}[t!]
\centering
\tableFont
\captionsetup{font=small}
\caption{\textbf{Experiments on Hypersim.} We observe that our method consistently outperforms the baseline, as well as the ManhattanDF. This is very interesting since, unlike ManhattanDF, we do not use any additional labels during training. Finally, we see that it does not matter for our method whether the MF is known beforehand or not. Therefore, the additional knowledge of MF is neither necessary nor helpful.}
\vspace{-2mm}
\begin{tabular}{|l|l|c|c|c|c|c|c|c|c|}
\hline
\multicolumn{2}{|c|}{} & PSNR$\uparrow$ & SSIM$\uparrow$ & Normals$\degree\downarrow$ & Pitch$\degree\downarrow$ & Roll$\degree\downarrow$ & Yaw$\degree\downarrow$ & D-MAE$\downarrow$&  D-RMSE$\downarrow$ \\ \hline \hline
\multirow{4}{*}{\makecell{Scenes \\ A}}
 & InstantNGP~\cite{mueller2022instant} (baseline)  
 & 25.86 & 0.871 & 57.12 & 6.18 & 6.46 & 19.25 & 0.064 & 0.102 \\ \cline{2-10}
 & ManhattanDF~\cite{guo2022neural} 
 & 26.51 & 0.868 & 40.69 & 1.23 & 1.01 & 5.08 & 0.053 & 0.087  \\ \cline{2-10}
 & Ours         
 & 27.20 & 0.864 & 37.30 & 0.40 & 0.50 & 0.52 & 0.053 & 0.093 \\ \cline{2-10}
 & Ours (MF known)  
 & 27.21 & 0.856 & 35.59 & 0.25 & 0.26 & 0.45 & 0.052 & 0.091 
 \\ \hline  \hline 
 
\multirow{4}{*}{\makecell{Scenes \\ B}}
 & InstantNGP~\cite{mueller2022instant}  (baseline)    
 & 20.75 & 0.811 & 60.12 & 6.06 & 7.92 & 15.87 & 0.105 & 0.151 \\ \cline{2-10}
 & ManhattanDF~\cite{guo2022neural}
 & 21.87 & 0.826 & 50.50 & 2.55 & 2.06 & 11.69 & 0.079 & 0.121 \\ \cline{2-10}
 & Ours         
 & 22.45 & 0.816 & 54.08 & 1.19 & 1.35 & 1.81 & 0.080 & 0.127 \\ \cline{2-10}
 & Ours (MF known)  
 & 22.51 & 0.813 & 50.59 & 0.51 & 0.65 & 0.55 & 0.078 & 0.126 
 \\ \hline  \hline 
 
\multirow{4}{*}{\makecell{Scenes \\ C}}
 & InstantNGP~\cite{mueller2022instant}   (baseline)   
 & 17.79 & 0.740 & 64.29 & 7.45 & 4.55 & 15.14 & 0.130 & 0.174 \\ \cline{2-10}
 & ManhattanDF~\cite{guo2022neural} 
 & 18.33 & 0.770 & 56.08 & 3.20 & 3.41 & 10.25 & 0.103 & 0.147  \\ \cline{2-10}
 & Ours         
 & 19.43 & 0.768 & 54.79 & 5.37 & 2.24 & 4.24 & 0.094 & 0.133 \\ \cline{2-10}
 & Ours (MF known)  
 & 19.29 & 0.764 & 55.12 & 3.64 & 4.03 & 9.48 & 0.094 & 0.135 
 \\ \hline \hline
 
 \multirow{3}{*}{\makecell{194 \\ scenes }}
 & InstantNGP~\cite{mueller2022instant}   (baseline)   
 & 20.47 & 0.783 & 61.34 & 6.56 & 6.99 & 21.75 & 0.104 & 0.146 \\ \cline{2-10}
 & ManhattanDF~\cite{guo2022neural}   
 & 20.94 & 0.794 & 52.81 & 1.72 & 2.32 & 13.48 & 0.097 & 0.140 \\ \cline{2-10}
 & Ours
 & 21.63 & 0.786 & 52.01 & 1.87 & 1.94 & 3.77 & 0.085 & 0.126  \\ \hline
\end{tabular}
\vspace{-3mm}
\label{table:hypersim}
\end{table*}


\noindent\textbf{Hypersim}~\cite{roberts2021hypersim} is a photorealistic synthetic dataset consisting of indoor scenes. It was created by leveraging a large repository of scenes created by professional artists, with 461 indoor scenes in total. This dataset is geometry-rich, containing a lot of details and lighting sources. Each scene has one or more camera trajectories available, where each trajectory has up to 100 views rendered in 1024×768. For each scene, camera calibration information is provided, as well as detailed per-pixel labels such as depth and surface normals. After cleaning up the scenes by discarding a few with insufficient camera views, as well as other problems, we were left with 435 scenes. 
We also kept only one camera trajectory per scene. We then evaluated the InstantNGP~\cite{mueller2022instant} baseline on all 435 scenes and made three divisions based on its PSNR on unseen views. 
\textbf{Hypersim-A} contains 20 scenes where the baseline was performing well above average, \textbf{Hypersim-B} contains 20 scenes where the baseline had near-average performance, and \textbf{Hypersim-C} contains 10 scenes where the baseline had a below-average performance. For each scene, we randomly assigned half views to the training split and the rest to the test split. 
We note that the test split often contains views that were partially or completely unobserved during training.
We also note that Hypersim is visually very realistic, geometry-rich, and challenging, considering it is synthetic.
This can be subjectively observed by inspecting rendered views.

\noindent\textbf{ScanNet}~\cite{dai2017scannet} is a real-world dataset consisting of indoor scenes. 
It was collected using a scalable RGB-D capture system that includes automated surface reconstruction and crowd-sourced semantic annotation.
The dataset contains 1613 indoor scenes, which are annotated with ground-truth camera poses, surface reconstructions, and instance-level semantic segmentations.
We use three scenes that were used in~\cite{guo2022neural}, where one-tenth of all views were uniformly sampled, leaving 303-477 views per scene. 
For each scene, the training and test split both contain half of the total views.

\noindent\textbf{Replica}~\cite{replica19arxiv} is a synthetic dataset featuring a diverse set of 18
indoor scenes. Each scene is equipped with photorealistic textures, allowing one to render realistic images from arbitrary camera poses. 
We use five scenes that were rendered and prepared in~\cite{zhi2021place}, where the rendered views also contain semantic segmentation labels. Each scene contains 900 views generated from random 6-DOF trajectories similar to hand-held camera motions in 640x480 resolution. We select 75 views for training and 75 for testing in each scene.
%
%

\subsection{Results}
\begin{figure}[t!]
    \vspace{1mm}
    \centering
    \captionsetup{font=small}
    \begin{subfigure}[b]{0.49\columnwidth}
        \centering
        \def\svgwidth{1\textwidth} 
\begingroup%
  \makeatletter%
  \providecommand\color[2][]{%
    \errmessage{(Inkscape) Color is used for the text in Inkscape, but the package 'color.sty' is not loaded}%
    \renewcommand\color[2][]{}%
  }%
  \providecommand\transparent[1]{%
    \errmessage{(Inkscape) Transparency is used (non-zero) for the text in Inkscape, but the package 'transparent.sty' is not loaded}%
    \renewcommand\transparent[1]{}%
  }%
  \providecommand\rotatebox[2]{#2}%
  \newcommand*\fsize{\dimexpr\f@size pt\relax}%
  \newcommand*\lineheight[1]{\fontsize{\fsize}{#1\fsize}\selectfont}%
  \ifx\svgwidth\undefined%
    \setlength{\unitlength}{500.38580947bp}%
    \ifx\svgscale\undefined%
      \relax%
    \else%
      \setlength{\unitlength}{\unitlength * \real{\svgscale}}%
    \fi%
  \else%
    \setlength{\unitlength}{\svgwidth}%
  \fi%
  \global\let\svgwidth\undefined%
  \global\let\svgscale\undefined%
  \makeatother%
  \begin{picture}(1,0.57995061)%
    \lineheight{1}%
    \setlength\tabcolsep{0pt}%
    \put(0,0){\includegraphics[width=\unitlength,page=1]{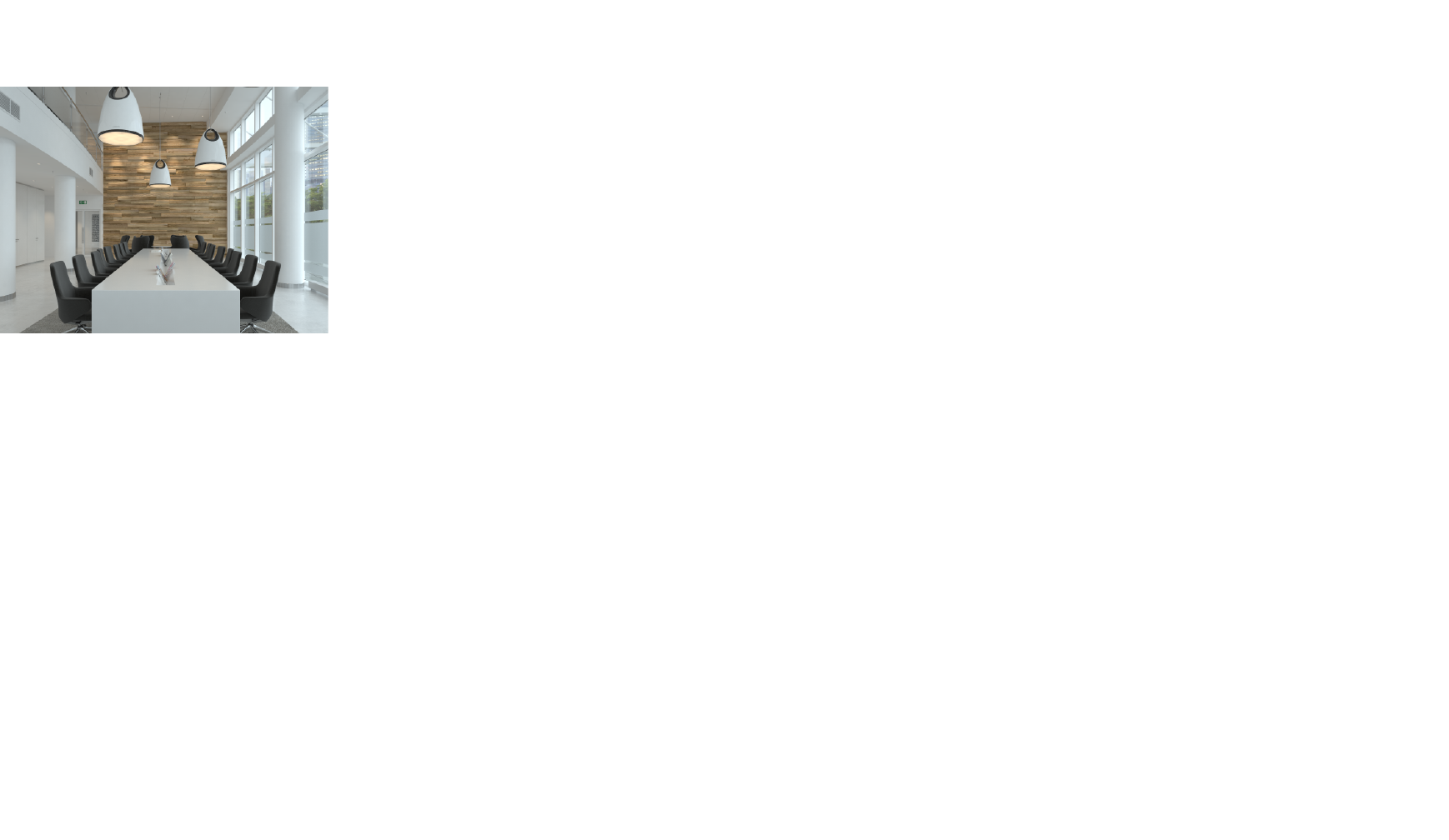}}%
    \put(-0.00299769,0.53198762){\color[rgb]{0,0,0}\makebox(0,0)[lt]{\lineheight{1.25}\smash{\begin{tabular}[t]{l}{\tiny Knowns}\end{tabular}}}}%
    \put(0,0){\includegraphics[width=\unitlength,page=2]{insNorm_c.pdf}}%
  \end{picture}%
\endgroup%

        \caption{InstantNGP~\cite{mueller2022instant} (baseline)  }
    \end{subfigure}
    \hfill    
    \begin{subfigure}[b]{0.49\columnwidth}
        \centering
        \def\svgwidth{1\textwidth} 
\begingroup%
  \makeatletter%
  \providecommand\color[2][]{%
    \errmessage{(Inkscape) Color is used for the text in Inkscape, but the package 'color.sty' is not loaded}%
    \renewcommand\color[2][]{}%
  }%
  \providecommand\transparent[1]{%
    \errmessage{(Inkscape) Transparency is used (non-zero) for the text in Inkscape, but the package 'transparent.sty' is not loaded}%
    \renewcommand\transparent[1]{}%
  }%
  \providecommand\rotatebox[2]{#2}%
  \newcommand*\fsize{\dimexpr\f@size pt\relax}%
  \newcommand*\lineheight[1]{\fontsize{\fsize}{#1\fsize}\selectfont}%
  \ifx\svgwidth\undefined%
    \setlength{\unitlength}{500.38580947bp}%
    \ifx\svgscale\undefined%
      \relax%
    \else%
      \setlength{\unitlength}{\unitlength * \real{\svgscale}}%
    \fi%
  \else%
    \setlength{\unitlength}{\svgwidth}%
  \fi%
  \global\let\svgwidth\undefined%
  \global\let\svgscale\undefined%
  \makeatother%
  \begin{picture}(1,0.57995061)%
    \lineheight{1}%
    \setlength\tabcolsep{0pt}%
    \put(0,0){\includegraphics[width=\unitlength,page=1]{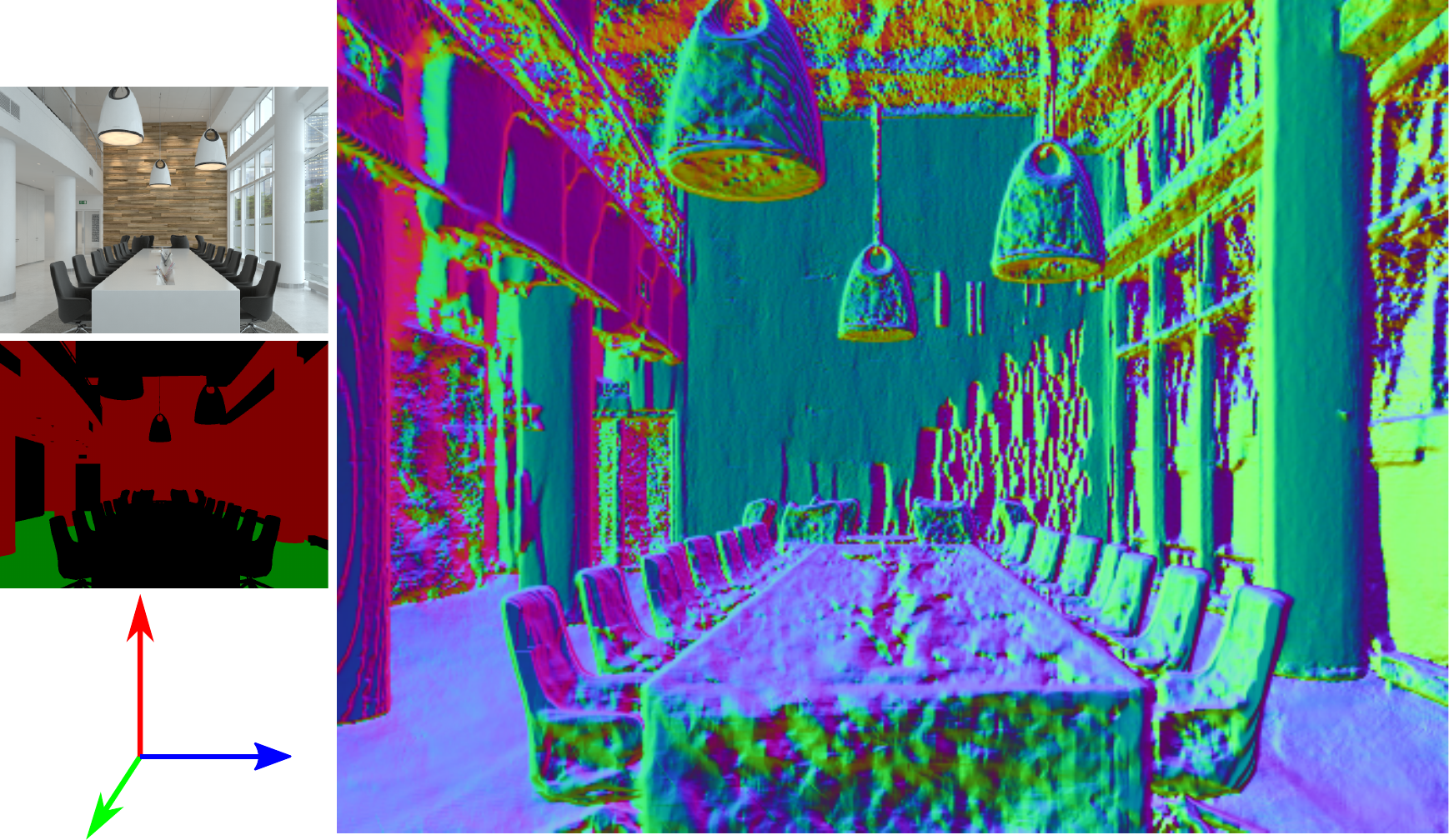}}%
    \put(-0.00299769,0.53198762){\color[rgb]{0,0,0}\makebox(0,0)[lt]{\lineheight{1.25}\smash{\begin{tabular}[t]{l}{\tiny Knowns}\end{tabular}}}}%
  \end{picture}%
\endgroup%

        \caption{ManhattanDF~\cite{guo2022neural}}
    \end{subfigure}
    \begin{subfigure}[b]{0.49\columnwidth}
        \centering
        \def\svgwidth{1\textwidth} 
\begingroup%
  \makeatletter%
  \providecommand\color[2][]{%
    \errmessage{(Inkscape) Color is used for the text in Inkscape, but the package 'color.sty' is not loaded}%
    \renewcommand\color[2][]{}%
  }%
  \providecommand\transparent[1]{%
    \errmessage{(Inkscape) Transparency is used (non-zero) for the text in Inkscape, but the package 'transparent.sty' is not loaded}%
    \renewcommand\transparent[1]{}%
  }%
  \providecommand\rotatebox[2]{#2}%
  \newcommand*\fsize{\dimexpr\f@size pt\relax}%
  \newcommand*\lineheight[1]{\fontsize{\fsize}{#1\fsize}\selectfont}%
  \ifx\svgwidth\undefined%
    \setlength{\unitlength}{500.38580947bp}%
    \ifx\svgscale\undefined%
      \relax%
    \else%
      \setlength{\unitlength}{\unitlength * \real{\svgscale}}%
    \fi%
  \else%
    \setlength{\unitlength}{\svgwidth}%
  \fi%
  \global\let\svgwidth\undefined%
  \global\let\svgscale\undefined%
  \makeatother%
  \begin{picture}(1,0.57995061)%
    \lineheight{1}%
    \setlength\tabcolsep{0pt}%
    \put(0,0){\includegraphics[width=\unitlength,page=1]{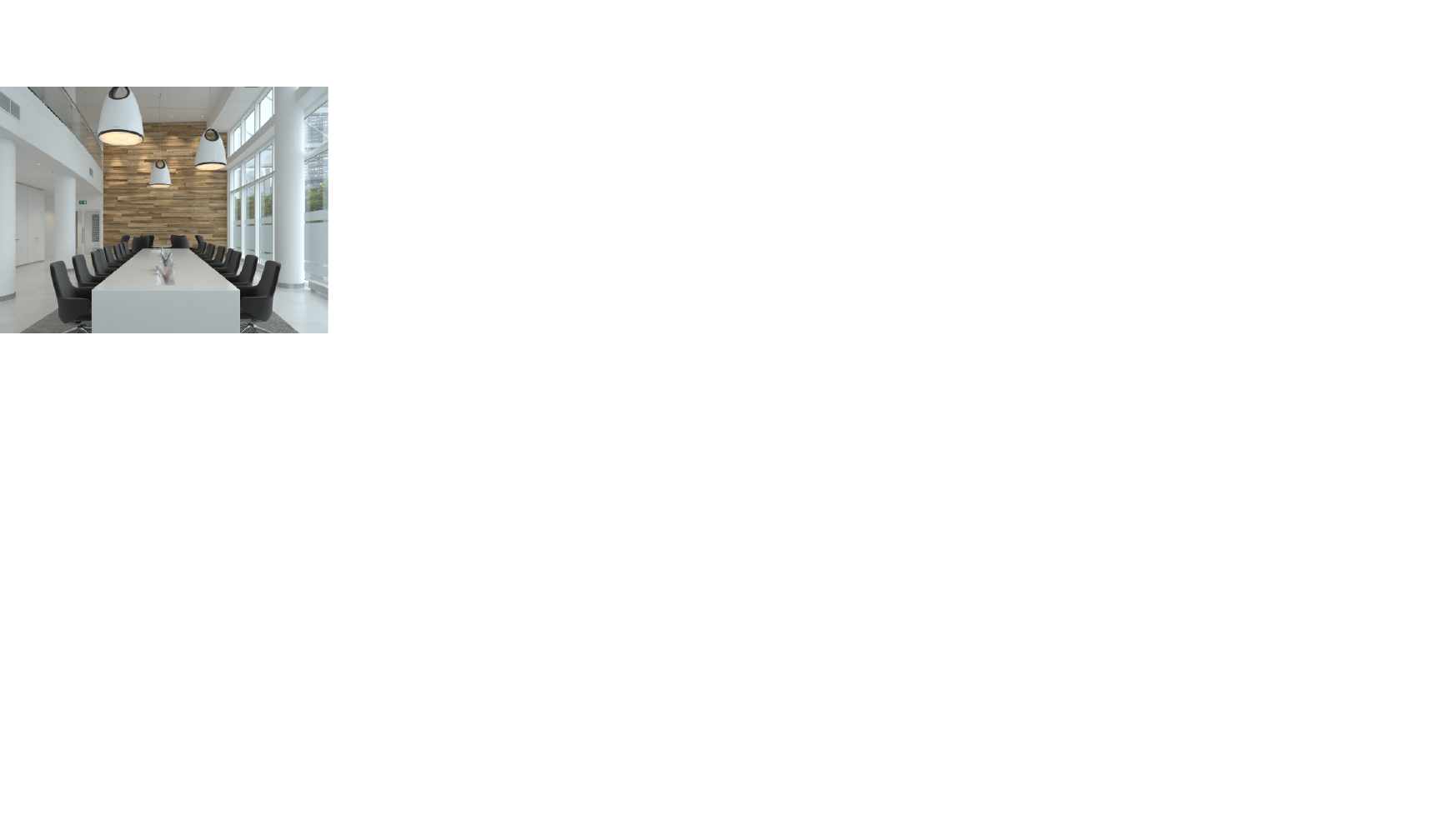}}%
    \put(-0.00299769,0.53198762){\color[rgb]{0,0,0}\makebox(0,0)[lt]{\lineheight{1.25}\smash{\begin{tabular}[t]{l}{\tiny Knowns}\end{tabular}}}}%
    \put(0,0){\includegraphics[width=\unitlength,page=2]{oursNorm_c.pdf}}%
  \end{picture}%
\endgroup%

        \caption{Ours}
    \end{subfigure}
    \hfill    
    \begin{subfigure}[b]{0.49\columnwidth}
        \centering
       \,\,\,\,\,\,\,\,\,\,\,\,\,\,\,\, \includegraphics[width=0.75\columnwidth]{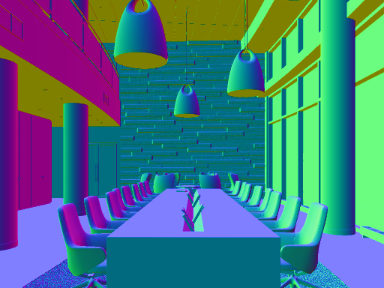}
        \caption{Grond truth}
    \end{subfigure}
    \vspace{-2mm}
    \caption{{\textbf{Leveraging the Manhattan prior (Hypersim-A).} Our method leverages the presence of many Manhattan objects and areas in the scene in a self-supervised fashion, which facilitates imposing geometrical constraints during learning. Our method offers plausible normals -- sometimes with missing details -- that help to better model the radiance fields. In contrast, ManhattanDF leverages the Manhattan prior through labels of only walls and floors. }}
    \label{fig:normals_vis}
    \vspace{-3mm}
\end{figure}

\noindent\textbf{Experiments on Hypersim:} We summarize the experiments performed on Hypersim in Table~\ref{table:hypersim}.
Our proposed method achieves clear improvements in comparison to the InstantNGP baseline, in terms of novel-view rendering, normals estimation, depth estimation, as well as MF recovery. This is consistent across different scene difficulties, namely splits A, B, and C.
Furthermore, our method also outperforms the ManhattanDF SotA. This is very interesting since, unlike ManhattanDF, we do not use any additional labels during training, other than the ground truth RGB. The ManhattanDF has partial access to ground truth MF axes, as well as wall and floor semantics, which directly implies having sparse surface normals ground truth. However, there are lots of Manhattan objects and areas in realistic scenes such as in Hypersim, other than the walls and floors. Our self-supervised method leverages the presence of many such objects and areas, to facilitate imposing geometrical constraints on the implicit representation during learning. This can be visually observed in Figure~\ref{fig:normals_vis}. 
Moreover, we observe that it does not matter for our method whether the MF is known beforehand, since our method recovers the MF in the clustering step automatically, and therefore similar performance is achieved in both cases. This is particularly exciting because our self-supervised method performs similarly to the supervised one (using the ground-truth MF). Finally, we show large-scale experiments on a larger set of 194 scenes, which lead to the same conclusions. We observed that the semantic loss of the ManhattanDF sometimes causes convergence issues on difficult scenes, which we partially alleviated with class weighting and label smoothing. Therefore, we report results on only 194 scenes where ManhattanDF converged. Additional results on all 435 scenes are provided in the supplementary.

\begin{figure*}[t!]
\centering
\addtolength{\tabcolsep}{-4.5pt}
\begin{tabular} {c||ccc||cc||cc|}
 \cline{2-8} & \multicolumn{3}{|c||}{\textbf{Hypersim}} & \multicolumn{2}{|c||}{\textbf{ScanNet}} & \multicolumn{2}{|c|}{\textbf{Replica}} \\ \cline{2-8}
 & RGB & Depth & Normals & RGB & Normals & RGB & Normals \\

 \multirow{1}{*}[35pt]{\rotatebox{90}{InstaNGP}} &
 \includegraphics[width=0.130\linewidth]{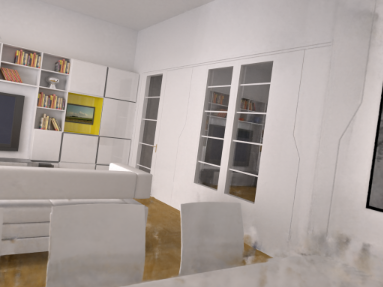}&
 \includegraphics[width=0.130\linewidth]{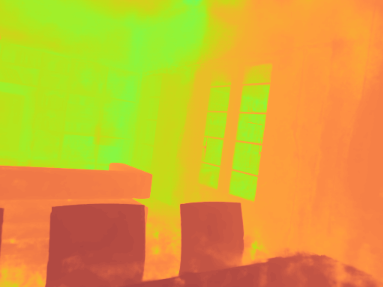}&
 \includegraphics[width=0.130\linewidth]{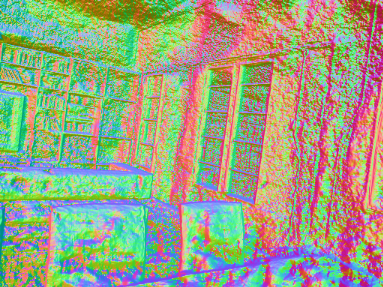}&
 \includegraphics[width=0.130\linewidth]{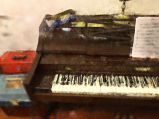}&
 \includegraphics[width=0.130\linewidth]{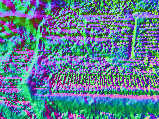}&
 \includegraphics[width=0.130\linewidth]{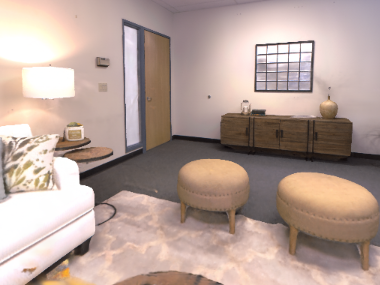}&
 \includegraphics[width=0.130\linewidth]{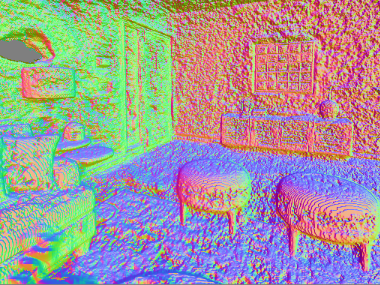} \\
 
 \multirow{1}{*}[32pt]{\rotatebox{90}{ManDF}} &
 \includegraphics[width=0.130\linewidth]{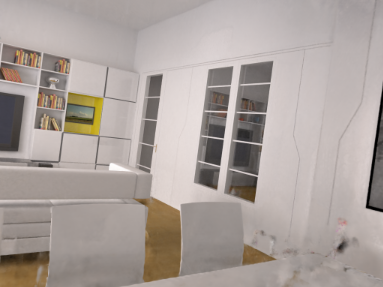}&
 \includegraphics[width=0.130\linewidth]{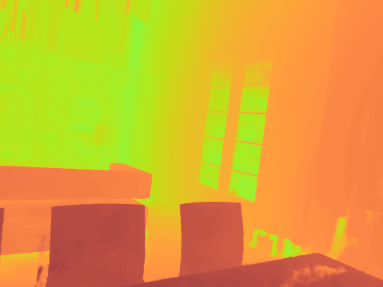}&
 \includegraphics[width=0.130\linewidth]{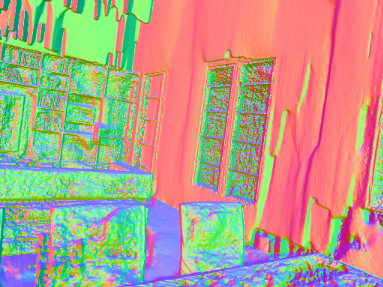}&
 \includegraphics[width=0.130\linewidth]{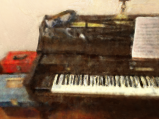}&
 \includegraphics[width=0.130\linewidth]{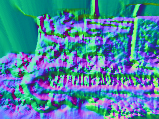}&
 \includegraphics[width=0.130\linewidth]{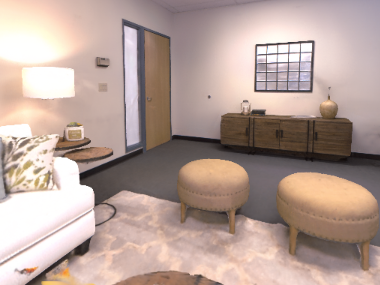}&
 \includegraphics[width=0.130\linewidth]{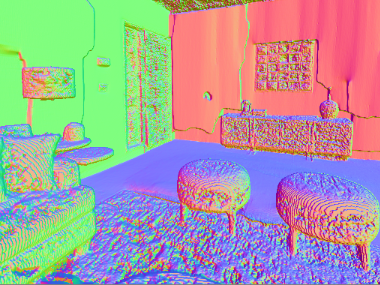} \\
 
 \multirow{1}{*}[25pt]{\rotatebox{90}{Ours}} &
 \includegraphics[width=0.130\linewidth]{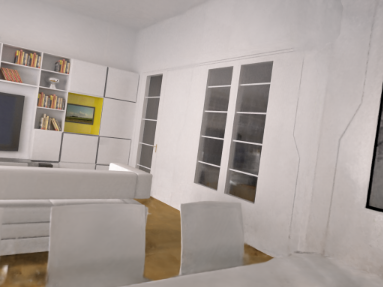}&
 \includegraphics[width=0.130\linewidth]{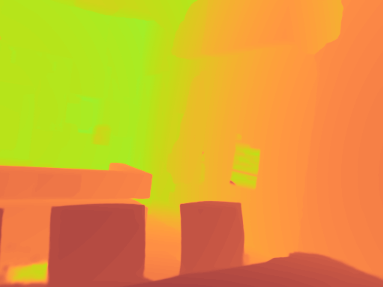}&
 \includegraphics[width=0.130\linewidth]{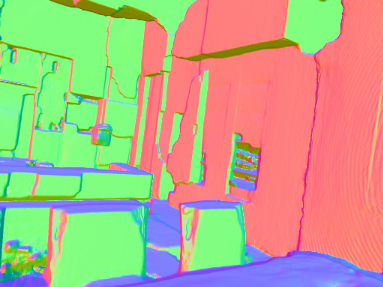}&
 \includegraphics[width=0.130\linewidth]{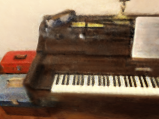}&
 \includegraphics[width=0.130\linewidth]{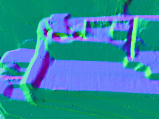}&
 \includegraphics[width=0.130\linewidth]{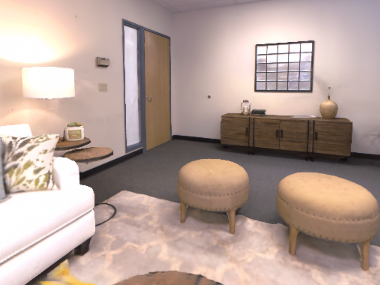}&
 \includegraphics[width=0.130\linewidth]{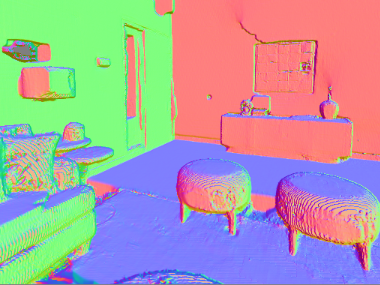} \\
 
 \multirow{1}{*}[20pt]{\rotatebox{90}{GT}} &
 \includegraphics[width=0.130\linewidth]{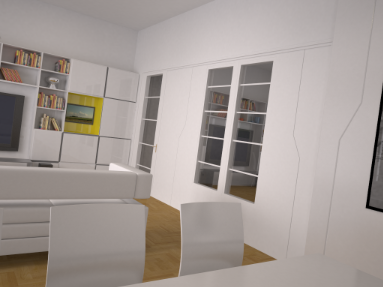}&
 \includegraphics[width=0.130\linewidth]{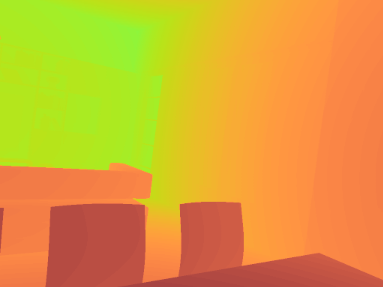}&
 \includegraphics[width=0.130\linewidth]{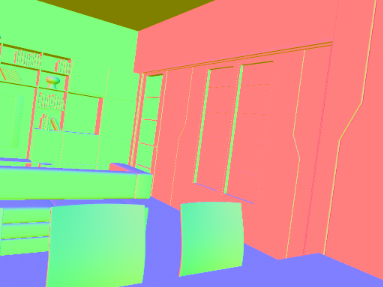}&
 \includegraphics[width=0.130\linewidth]{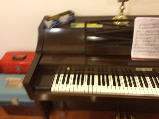}&
 \includegraphics[width=0.130\linewidth]{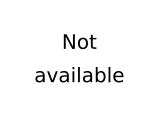}&
 \includegraphics[width=0.130\linewidth]{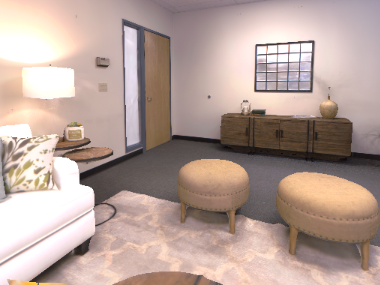}&
 \includegraphics[width=0.130\linewidth]{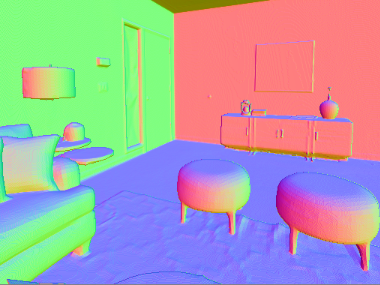} \\ 
\end{tabular}
\vspace{-3.5mm}  
\caption{\textbf{Qualitative results.} Our method leverages many Manhattan objects and surfaces in the scene, which improves the implicit geometrical representation compared to the baseline. Unlike ManhattanDF, our method relies on many cues other than the walls \& floors, which leads to a better representation of some objects (e.g. the white table in the bottom right corner of the example from columns 1-3). For more qualitative results, please refer to the supplementary material.}
\label{fig:big_vis}
\vspace{-2.5mm}  
\end{figure*}

\noindent\textbf{Experiments on ScanNet:} In Table~\ref{table:scannet_man}, we examine the behavior of our proposed method on real-world indoor scenes. Our method achieves clear improvements in comparison to the InstantNGP baseline, as well as to ManhattanDF, in terms of all measured metrics. Additionally, following the experimental setting of ManhattanDF~\cite{guo2022neural}, we train all methods with additional sparse depth supervision -- where the sparse depth is obtained from the SfM pipeline. 
Our proposed method is also superior in this setting.

\begin{table}[t!]
\centering
\tableFont
\captionsetup{font=small}
 \vspace{-2mm}  
\caption{\textbf{Experiments on real-world ScanNet data.} 
Our method outperforms both the baseline and ManhattanDF. This is the case both with and without supervising with sparse depth from SfM.
}
 \vspace{-2mm}  
\begin{tabular}{|c|c|c|c|}
\hline
 & PSNR$\uparrow$  & SSIM $\uparrow$ & Depth$\downarrow$-MAE \\ \hline \hline
 InstantNGP~\cite{mueller2022instant} (baseline)
& 17.78 & 0.587 & 0.119  \\  \hline
RegNeRF~\cite{niemeyer2022regnerf}
& 18.73 & 0.618 & 0.102 \\ \hline
ManhattanDF~\cite{guo2022neural}
& 18.68 & 0.614 & 0.112  \\  \hline
Ours
& 20.79 & 0.643 & 0.072  \\  \hline \hline 
\multicolumn{4}{|c|}{+ additional sparse depth from SfM} \\ \hline
InstantNGP~\cite{mueller2022instant} (baseline)
& 20.70 & 0.631 & 0.048  \\  \hline
ManhattanDF~\cite{guo2022neural} 
& 21.53 & 0.640 & 0.052  \\  \hline 
Ours 
& 22.25 & 0.667 & 0.033  \\  \hline


\end{tabular}
 \vspace{-2mm}  
\label{table:scannet_man}
\end{table}

\begin{table}[t!]
\centering
\tableFont
\captionsetup{font=small}
\vspace{-1mm}  
\caption{\textbf{Experiments on Replica.} Our method outperforms the baseline, and it shows similar performance as ManhattanDF, which leverages additional labels during training.}
 \vspace{-2mm}  
\begin{tabular}{|c|c|c|c|}
\hline
 & PSNR$\uparrow$  & SSIM $\uparrow$ & Depth$\downarrow$-MAE \\ \hline \hline
InstantNGP~\cite{mueller2022instant}  (baseline) 
& 34.30 & 0.944 & 0.022  \\  \hline
Semantic-NeRF~\cite{zhi2021place}
& 34.08 & 0.938 & 0.014 \\ \hline
ManhattanDF~\cite{guo2022neural} 
& 35.24 & 0.944 & 0.008  \\  \hline
Ours 
& 35.13 & 0.944 & 0.011  \\  \hline

\end{tabular}
\label{table:replica_2}
\vspace{-4mm}  
\end{table}

\noindent\textbf{Experiments on Replica:} 
We summarize experiments performed on the Replica dataset in Table~\ref{table:replica_2}. It can be observed that the InstantNGP baseline already performs very well on this dataset, leaving not much room for further improvements. Therefore, we consider Replica as an easier dataset. Nevertheless, our method still performs better than the baseline. As expected, ManhattanDF also improves the baseline similarly. Recall that the ManhattanDF uses additional labels for supervision.  
We also note that Replica has noticeably more walls and floors, and less of other objects, compared to other more complex datasets.


\begin{figure}[t!]
    \vspace{-3mm}  
    \centering
    \captionsetup{font=small}
    \includegraphics[width=0.93\columnwidth]{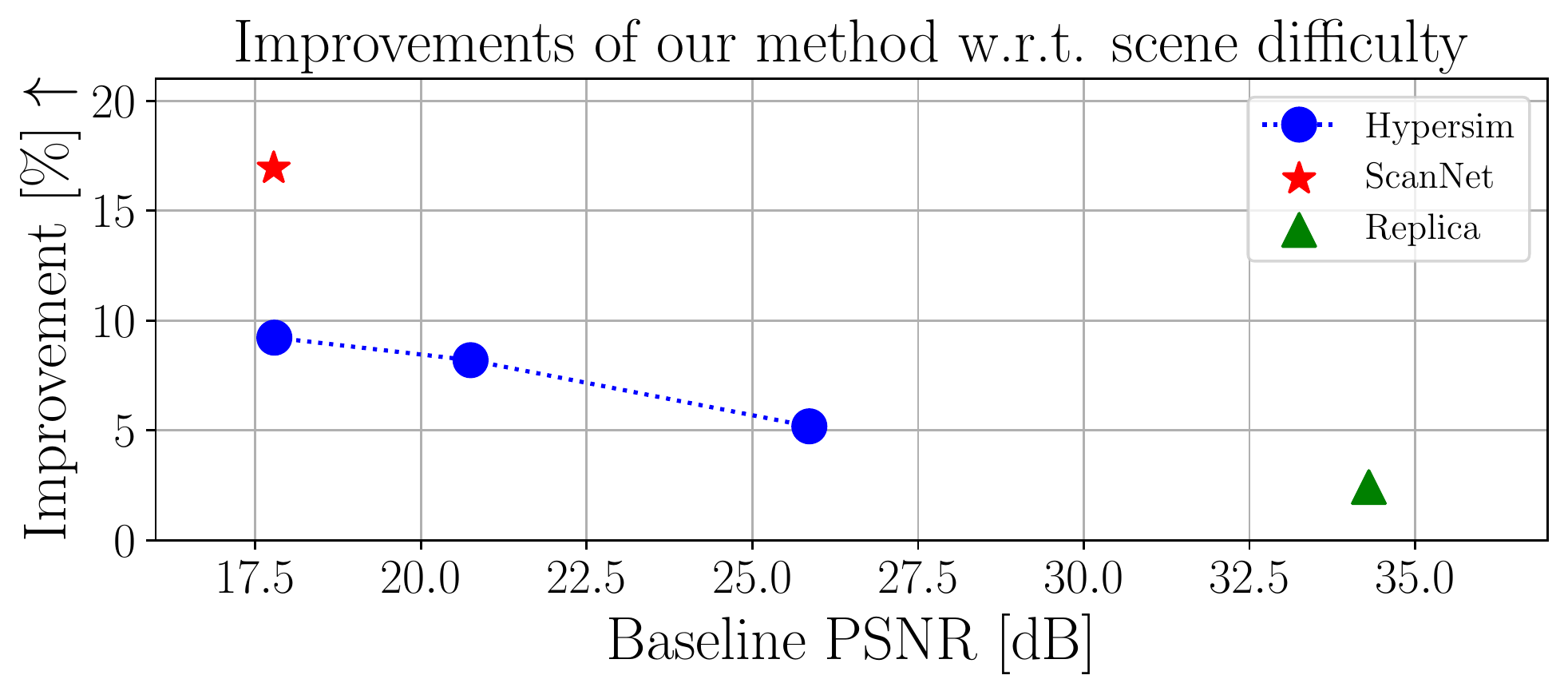}
    \vspace{-4mm}  
    \caption{{\textbf{Improvements vs. scene difficulty.} Our method achieves the biggest improvements on scenes of hard and moderate difficulty, by leveraging many Manhattan objects and surfaces.}}
    \label{fig:improvement_wrt_difficulty}
    \vspace{-3mm}  
\end{figure}

\noindent\textbf{Qualitative results:} We depict the visual results of the discussed experiments in Figure~\ref{fig:big_vis}. Our method leverages many Manhattan objects and surfaces in the scene, which improves the geometrical structure of 3D compared to the InstantNGP baseline. This is visible in surface normals and depth, obtained using volume rendering. Furthermore, we observe that, unlike ManhattanDF, our method relies on many Manhattan cues other than the walls \& floors. This leads to a better representation of some such objects, e.g. the white table in the bottom right corner of the example from columns 1-3. Moreover, we observe that our method is able to cope with difficult scenes and views, where other methods struggle. For more qualitative results, please refer to the supplementary material.

\noindent\textbf{Improvements with respect to scene difficulty:}
We analyze the improvements by our method for different scene difficulties. We decided on scene difficulty based on the novel-view rendering performance of the InstantNGP baseline. 
In Figure~\ref{fig:improvement_wrt_difficulty}, we see that our method brings the most benefits for scenes of hard and moderate difficulties, thanks to the Manhattan scene prior. 

\noindent\textbf{Sparse training views:}
In order to gain more insight, we examine the behavior of training neural radiance fields with sparse training input views. The results on the Hypersim-A dataset are presented in Table~\ref{table:sparse_view}.
Our proposed method clearly outperforms the InstantNGP baseline, as well as ManhattanDF,  when trained with 6, 9, and 12 input views.
\begin{table}[t!]
\vspace{-3mm}  
\centering
\tableFont
\captionsetup{font=small}
\caption{\textbf{Sparse training views.} Our proposed method clearly exhibits the best performance, for all cases of input view sparsity.}
\vspace{-3mm}  

\begin{tabular}{|c|c|c|c|c|}
\hline
\multicolumn{2}{|c|}{} & PSNR$\uparrow$  & SSIM $\uparrow$ & Depth$\downarrow$-MAE \\ \hline \hline
\multirow{3}{*}{\makecell{12 \\ views}} 
& InstantNGP~\cite{mueller2022instant}
& 18.02 & 0.706 & 0.138  \\  \cline{2-5}
& ManhattanDF~\cite{guo2022neural} 
& 19.45 & 0.750 & 0.116  \\  \cline{2-5} 
& Ours 
& 20.50 & 0.760 & 0.104  \\  \hline \hline
\multirow{3}{*}{\makecell{9 \\ views}} 
& InstantNGP~\cite{mueller2022instant}
& 16.79 & 0.661 & 0.154  \\  \cline{2-5}
& ManhattanDF~\cite{guo2022neural} 
& 18.04 & 0.714 & 0.130  \\  \cline{2-5} 
& Ours 
& 19.14 & 0.728 & 0.120  \\  \hline \hline
\multirow{3}{*}{\makecell{6 \\ views}} 
& InstantNGP~\cite{mueller2022instant}
& 15.75 & 0.582 & 0.178  \\  \cline{2-5}
& ManhattanDF~\cite{guo2022neural}
& 16.00 & 0.639 & 0.159  \\  \cline{2-5}
& Ours
& 16.67 & 0.667 & 0.158  \\  \hline 

\end{tabular}
\vspace{-2mm}  
\label{table:sparse_view}
\end{table}

\begin{figure}[t!]
    \centering
    \captionsetup{font=small}
    \begin{subfigure}[b]{\columnwidth}
        \centering
        \includegraphics[width=0.8\columnwidth]{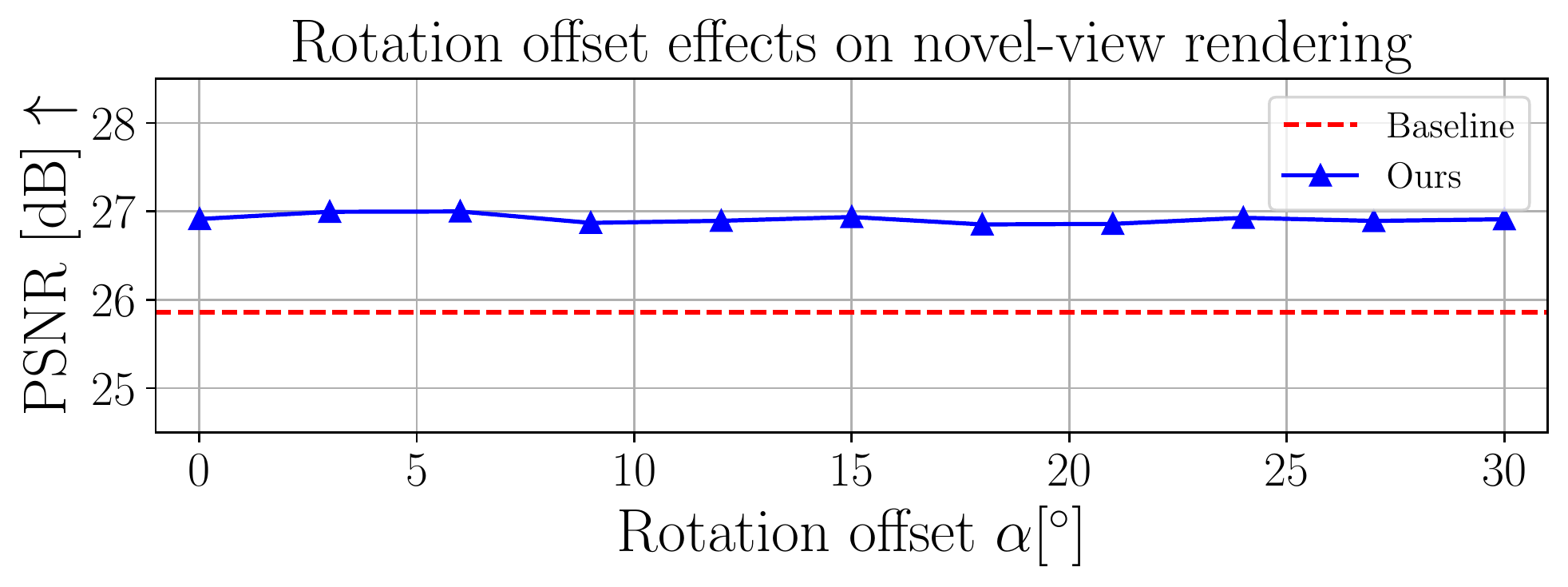}
        \vspace{-2.5mm}   
        \caption{Novel-view rendering quality remains robust to the rotation offset $\alpha$.}
        \label{fig:rotation_offset_psnr}
    \end{subfigure}
    \begin{subfigure}[b]{\columnwidth}
        \centering
        \includegraphics[width=0.8\columnwidth]{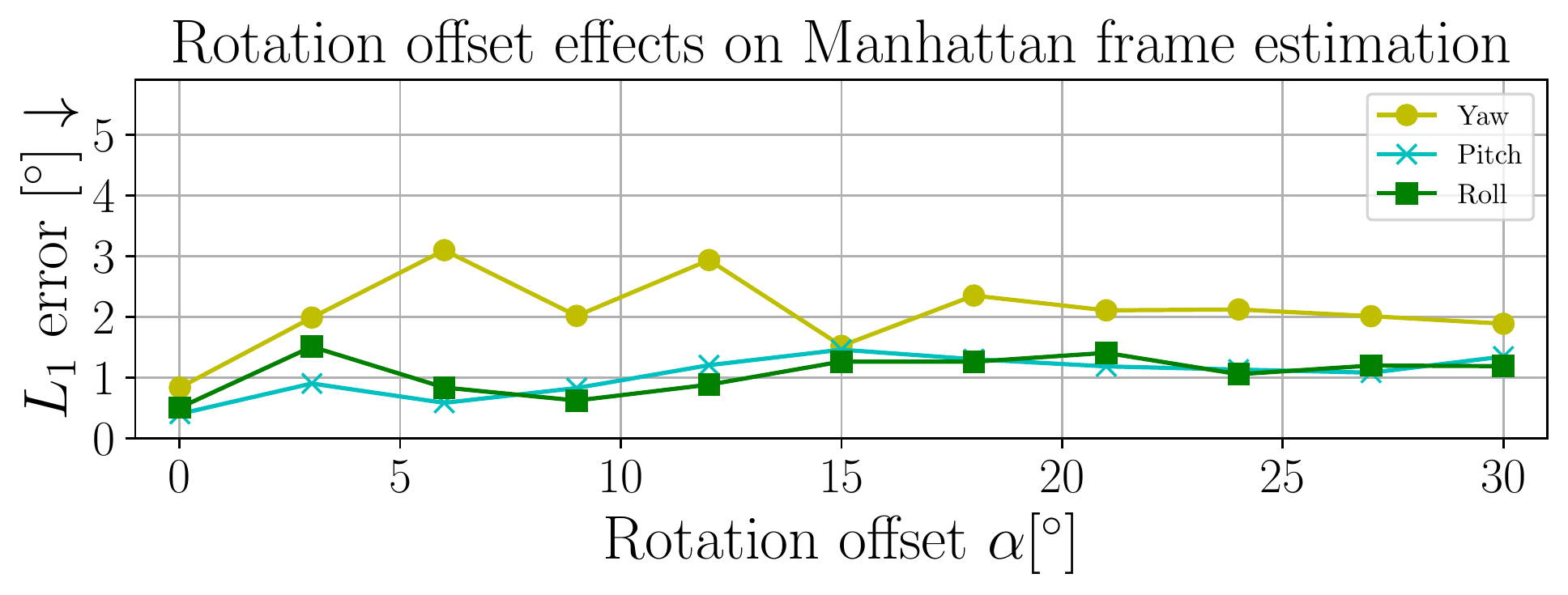}
        \vspace{-2.5mm}   
        \caption{MF estimation remains robust to the rotation offset $\alpha$.}
        \label{fig:rotation_offset_angles}
    \end{subfigure}
    \vspace{-6mm}   
    \caption{{\textbf{Finding the MF.} We test the robustness of our method by introducing the rotation offsets on the canonical MF.}}
    \label{fig:rotation_offset}
    \vspace{-3mm}  
\end{figure}

\noindent\textbf{Finding the MF:} 
We test the robustness of our proposed method for finding the Manhattan frame, by introducing a simultaneous offset $\alpha$ in the yaw, pitch, and roll on the canonical MF. The experiments are performed on Hypersim A and can be found in Figure~\ref{fig:rotation_offset}. In Figure~\ref{fig:rotation_offset_psnr} we observe that the novel-view rendering quality remains largely robust to the rotation offset $\alpha$. 
We note that we increase the scene bounding box by the same factor for all experiments in Figure~\ref{fig:rotation_offset}, to make sure that objects remain within the voxel grid for maximal $\alpha$. This slightly decreases the resolution of grid element representations, so PSNR is slightly lower than in Table~\ref{table:hypersim}.
Furthermore, in Figure~\ref{fig:rotation_offset_angles} we see that the MF estimation remains robust to the rotation offset.

\begin{table}[t!]
\centering
\tableFont
\captionsetup{font=small}
\caption{\textbf{Ablation study on Hypersim-A.} Both of our proposed loss terms contribute to the overall performance.}
 \vspace{-3mm}  
\begin{tabular}{|c|c|c|c|c|c|c|}
\hline
 & PSNR$\uparrow$  & Norm.$\degree\downarrow$ & Depth$\downarrow$&  Rot.$\degree\downarrow$ \\ \hline \hline
Only $\mathcal{L}_{img}$
& 25.86 & 57.12 & 0.064 & 10.63 \\  \hline \hline
+ $\mathcal{L}_{ort}$  
& 27.21 & 50.07 & 0.058 &  5.39 \\  \hline
+ $\mathcal{L}_{ctr}$     
& 27.06 & 36.09 & 0.053 &  0.57\\  \hline
+ $\mathcal{L}_{ort}$ + $\mathcal{L}_{ctr}$ (Ours)
& 27.20 & 37.30 & 0.053 &  0.47 \\ \hline 
Ours + MF known 
& 27.21 & 35.59 & 0.052 &  0.32 \\  \hline \hline

Ours (no delay)
& 27.06 & 35.78 & 0.054 &  0.39 \\ \hline  
Ours (no w lin.)
& 27.01 & 37.86 & 0.055 &  0.60 \\  \hline

\end{tabular}
 \vspace{-4mm}  
\label{table:ablation}
\end{table}

\noindent\textbf{Ablation study:} We report our ablation study in Table~\ref{table:ablation}. Both of our proposed losses contribute to the overall performance. Furthermore, turning on $\mathcal{L}_{ort}$ and $\mathcal{L}_{centr}$ after $500$ steps, and linearly increasing their weight also helps slightly. Please refer to the supplementary for more details.

\section{Conclusion}
We demonstrated the possibility of exploiting the Manhattan scene prior without needing any additional supervision. This is achieved by performing robust clustering of explicit normals, followed by the search of the Manhattan frame (MF) whose existence is based on the assumed prior. The sought MF is obtained from the orthogonal clusters, which are later used to self-supervise the neural representation learning. Our self-supervision encourages the normals of Manhattan surfaces to group into three orthogonal directions. Our experiments on three indoor datasets demonstrate that the proposed method not only benefits from building parts (such as walls and floors) but also exploits many other Manhattan scene parts (such as furniture). Both quantitative and qualitative evaluations reveal the benefit of the proposed method in terms of, improved performance over the established baselines and competitive results to state-of-the-art methods that use additional labels for supervision.  Our method has the potential to be extended in other higher expressiveness scene priors, such as Atlanta world and the mixture of Manhattan frames.

\noindent \textbf{Limitations:} One limitation of our method is that it sometimes produces surface normals with missing details or ``blocky" artifacts. Nevertheless, this usually offers better novel-view RGB rendering, compared to not imposing any structure priors. Another limitation is that our method is not beneficial for very easy scenes. For a detailed discussion of limitations, please refer to the supplementary material.

\section*{Acknowledgments}
This research was co-financed by Innosuisse under the project Know Where To Look, Grant 59189.1 IP-ICT, and
partially funded by the Ministry of Education and Science of Bulgaria (support for INSAIT, part of the Bulgarian National Roadmap for Research Infrastructure). Also, Carlos Oliveira helped with splendid visual renderings.

\appendix

\section{Supplementary Overview}
This supplementary material provides additional details, which complement the main paper. 
We first give additional details about our implementation and experimental setup in Section~\ref{sec:supp:implementation_details}.
In Section~\ref{sec:supp:further_analysis}, we complement the results from the main paper by showing additional comparisons and studies.
Furthermore, Section~\ref{sec:supp:limitations} comments about the limitations and failures of our approach.
Finally, we provide additional qualitative examples in Section~\ref{sec:supp:qualitative_results}.

\section{Implementation Details}
\label{sec:supp:implementation_details}
In this paragraph we describe the experimental setup when using  the InstantNGP backbone. We use a hash table of size $2^{19}$ containing $16$ levels with $2$ features per level, maximum grid resolution of $2048$, and an occupancy grid of resolution $128$. We jointly train the neural network weights and the hash table entries by applying Adam with $\epsilon=10^{-15}$ and a learning rate of $10^{-2}$. We also apply $\mathcal{L}_2$ regularization with a factor of $10^{-6}$, but only on neural network weights. These choices are based on suggestions in~\cite{mueller2022instant}. Also, for the sake of efficiency, we update the density grid after every 16 steps similarly to the procedure described in~\cite{mueller2022instant}.
To further stabilize the training we also use the opacity regularization~\cite{Lombardi_2019} with a factor of $10^{-3}$, as well as gradient norm clipping with a factor of $0.05$. We also use a cosine annealing learning rate scheduler and perform each training on $30$k iterations with batch size $8190$. 

When it comes to our proposed method we set the loss weights $\lambda_{ctr}=\lambda_{ort} = 2\cdot10^{-3}$ in the case of Hypersim, $1\cdot10^{-2}$ in the case of ScanNet, and $5\cdot10^{-4}$ in the case of Replica.
We turn on $\mathcal{L}_{ort}$ and $\mathcal{L}_{ctr}$ after $500$ steps, and linearly increase their $\lambda$ weights to the specified values over the next $2500$ steps. Also, when using methods which rely on explicit surface normals, we randomly sample rays for one-third of every batch size and select their left and upper pixel neighbor to form a triplet to facilitate obtaining these explicit normals. We call this a triplet triangle. Every triangle in the batch is sampled randomly from a set of all possible triangle triplets of all available images.

For k-means clustering, we use $k=20$ clusters when processing training batches. However, in order to estimate the Manhattan frame for all methods after the training ends, we cluster the normals from the whole test set into $k=30$ clusters. In addition, during both training and testing, we merge the three selected orthogonal clusters with their opposites. 
This is achieved by comparing the similarity of a selected cluster centroid $n_i$ and the opposite vector of every other cluster centroid $(-\mathsf{c}_j)$. In the case $|\mathsf{n}_i^{\intercal} (-\mathsf{c}_j)| > 1 - t$ holds true, all cluster elements of $\mathcal{C}_j$ are multiplied by $-1$ and added to $\mathcal{N}_i$. Also, if any cluster centroid $\mathsf{c}_j$ is close to one of the selected centroids $n_i$ ($|\mathsf{n}_i^{\intercal} \mathsf{c}_j| > 1 - t)$, elements from $\mathcal{C}_j$ are added to $\mathcal{N}_i$.

During the implementation of ManhattanDF~\cite{guo2022neural}, we initially had stability issues with the semantic segmentation cross-entropy loss. In order to alleviate these issues, we used label smoothing with a factor of $0.1$, as well as a class weight of $0.3$ for the background class. Also, in the Hypersim dataset, the scenes are much richer in content, and there are less wall \& floor labels compared to Replica. Therefore we merged the floor class with the floor mat class, and we also merged the window class with the wall class. We also observed in Hypersim that in a small number of scenes the ceiling is labeled as wall, or that there are no wall \& floor labels since they are all labeled as void. 

Also, when it comes to comparing baselines and different methods on different datasets, we always perform a thorough search over additional hyperparameters before reporting results. 
On Hypersim the weight for the semantic loss was $1 \cdot 10^{-1}$, while the weight for surface normals loss was $5 \cdot 10^{-4}$ in the regular case.
For ScanNet, the weight for the semantic loss was $1 \cdot 10^{-1}$, while the weight for surface normals loss was $5 \cdot 10^{-3}$.
When using additional sparse depth (from SfM) supervision for ScanNet, the depth loss weight was $1 \cdot 10^{-1}$, while the surface normals loss weight was changed to $1 \cdot 10^{-2}$.
For Replica, the weight for the semantic loss was $4 \cdot 10^{-2}$, while the weight for surface normals loss was $5 \cdot 10^{-4}$.

\section{Further Analysis}
\label{sec:supp:further_analysis}
In this section, we perform further analyses of the experiments presented in \ifx\FileIsMerged\undefined
Section 5 of the main paper.
\else
Section~\ref{sec:Experiments} of the main paper.
\fi

\begin{table*}[t!]
\centering
\tableFont
\captionsetup{font=small}
\caption{\textbf{Experiments on all Hypersim scenes.} We observe that our method outperforms the baseline  with respect to all observed metrics.}
\begin{tabular}{|l|l|c|c|c|c|c|c|c|c|}
\hline
\multicolumn{2}{|c|}{} & PSNR$\uparrow$ & SSIM$\uparrow$ & Normals$\degree\downarrow$ & Pitch$\degree\downarrow$ & Roll$\degree\downarrow$ & Yaw$\degree\downarrow$ & D-MAE$\downarrow$&  D-RMSE$\downarrow$ \\ \hline \hline
 \multirow{2}{*}{\makecell{435 \\ scenes }}
 & InstantNGP~\cite{mueller2022instant}   (baseline)   
 & 19.17 & 0.729 & 61.88 & 7.12 & 7.08 & 21.99 & 0.119 & 0.162 \\ \cline{2-10}
 & Ours
 & 20.17 & 0.737 & 54.93 & 3.15 & 3.05 & 8.17 & 0.100 & 0.142  \\ \hline
\end{tabular}
\vspace{-3mm}
\label{table:supp:hypersim_all}
\end{table*}

 \noindent \textbf{Results in 435 scenes of Hypersim:} In Table~\ref{table:supp:hypersim_all} we present results on all Hypersim scenes. Our proposed method performs better than the baseline with respect to all observed metrics. We were not able to evaluate ManhattanDF on all scenes, because experiments on a large portion of the scenes had convergence issues related to the specific loss function.

\begin{figure}[t!]
    \centering
    \captionsetup{font=small}
    \begin{subfigure}[b]{\columnwidth}
        \centering
        \includegraphics[width=\columnwidth]{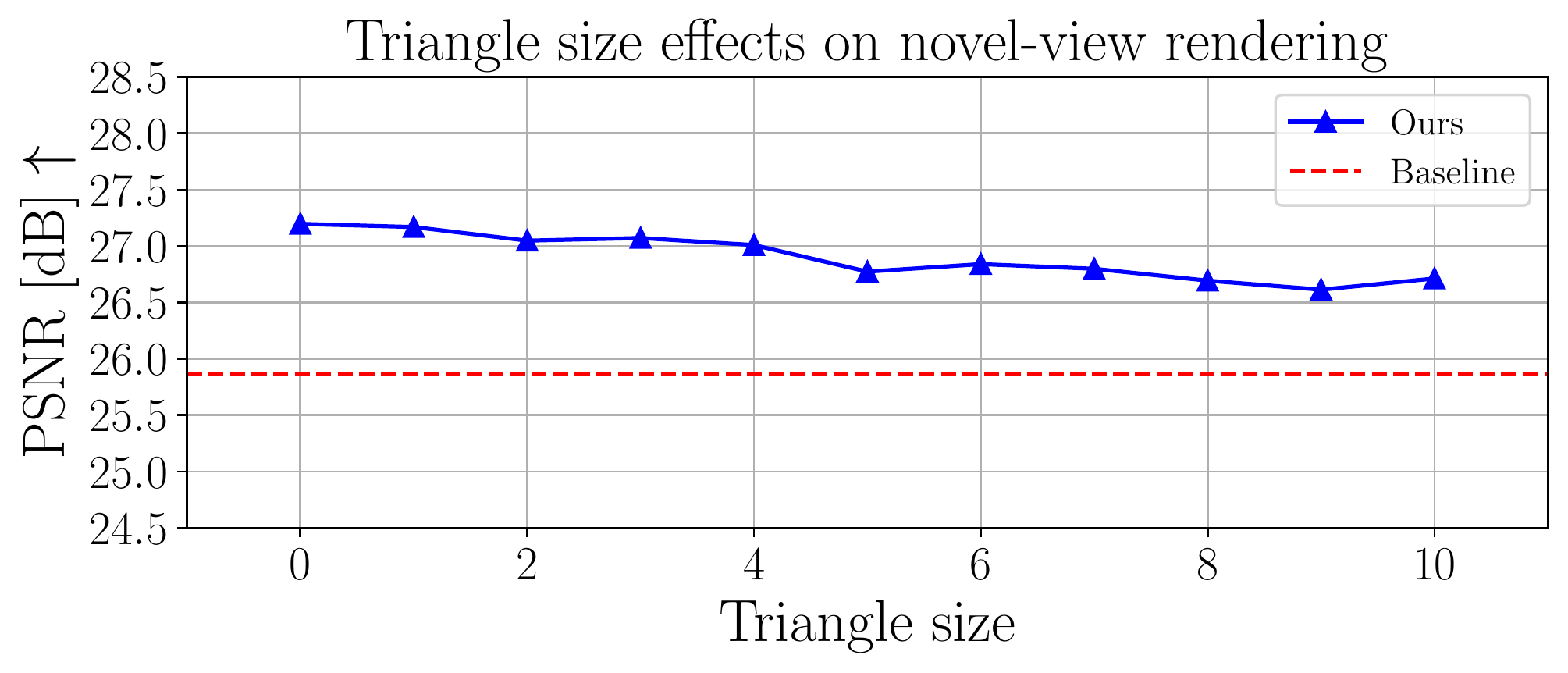}
        \caption{Triangle size choice effects on novel-view rendering.}
        \label{fig:supp:triangle_size_psnr}
    \end{subfigure}
    \begin{subfigure}[b]{\columnwidth}
        \centering
        \includegraphics[width=\columnwidth]{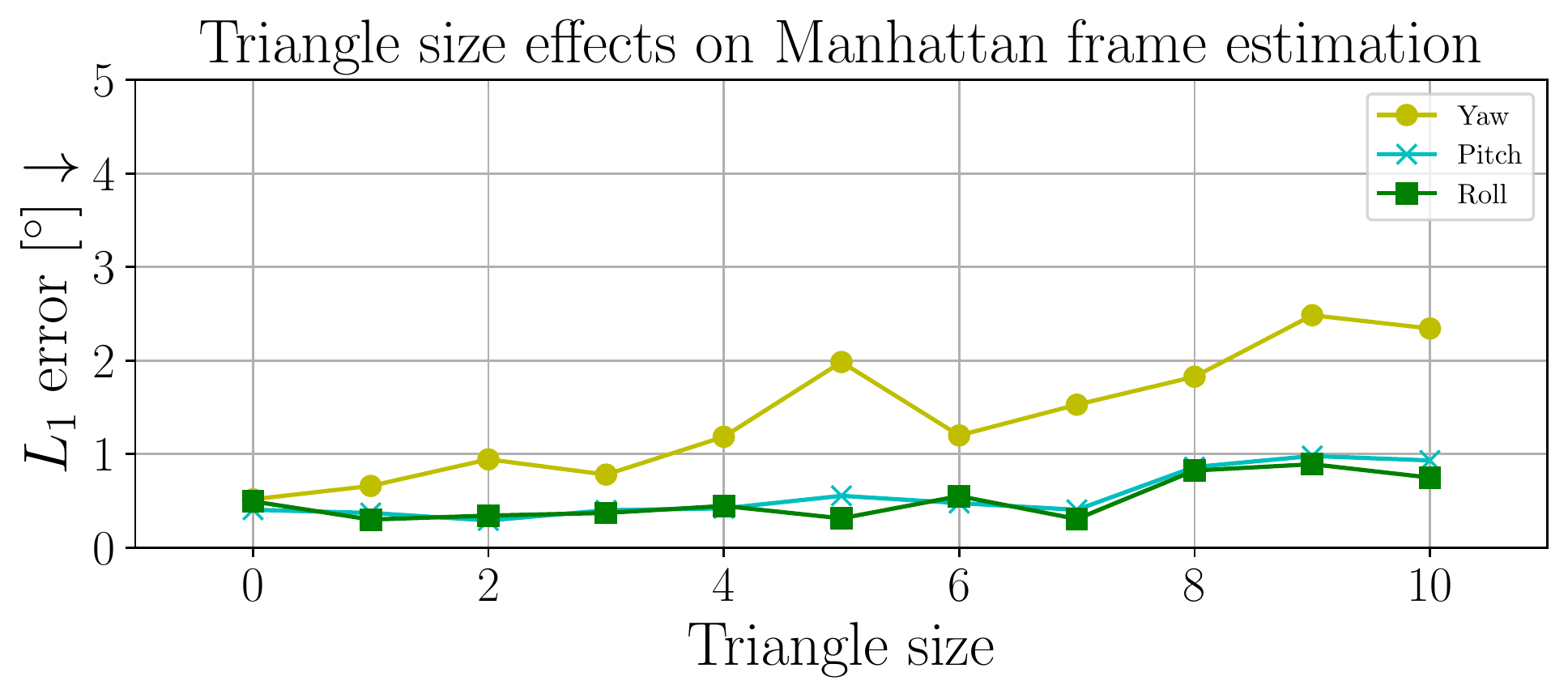}
        \caption{Triangle size choice effects on MF estimation.}
        \label{fig:supp:triangle_size_angles}
    \end{subfigure}
    \caption{{\textbf{Triplet triangle size choice effects on Hypersim-A.} The performance of our method slightly decreases with the triangle size increase. This is due to a bigger probability that the triplet will not lie on a planar surface segment, and thus the estimated normals will contain certain error. }}
    \label{fig:supp:triangle_size}
\end{figure}
\noindent \textbf{Triplet triangle size:} As previously mentioned, when using methods that rely on explicit surface normals, we randomly sample rays for one-third of every batch size and select their left and upper pixel neighbor to form a triplet triangle to facilitate obtaining explicit normals. We call this a triangle of size $0$, since the immediate left and upper neighbor are taken to form a triplet. Similarly, a triangle of size $k$ is when there is a $k$ pixel gap between the selected pixel and it's left and upper triplet pair. In Figure~\ref{fig:supp:triangle_size} we show results of our method for different triangle sizes. As the triangle size increases, the performance of our method slightly decreases in terms of novel-view rendering as well as MF estimation. With larger triangle sizes, there is a bigger probability that the triplet will not lie on a planar surface segment and thus normals estimated with a planar assumption will contain a certain degree of error. Nevertheless, this study shows that whenever one is certain about the bigger planar regions, our method of explicit normal computation can successfully be applied to potentially gain both efficiency and performance.  The further study in this regard however is out of the scope of this work, which remains as our future work. 

\begin{figure}[t!]
    \centering
    \captionsetup{font=small}
    \begin{subfigure}[b]{\columnwidth}
        \centering
        \includegraphics[width=\columnwidth]{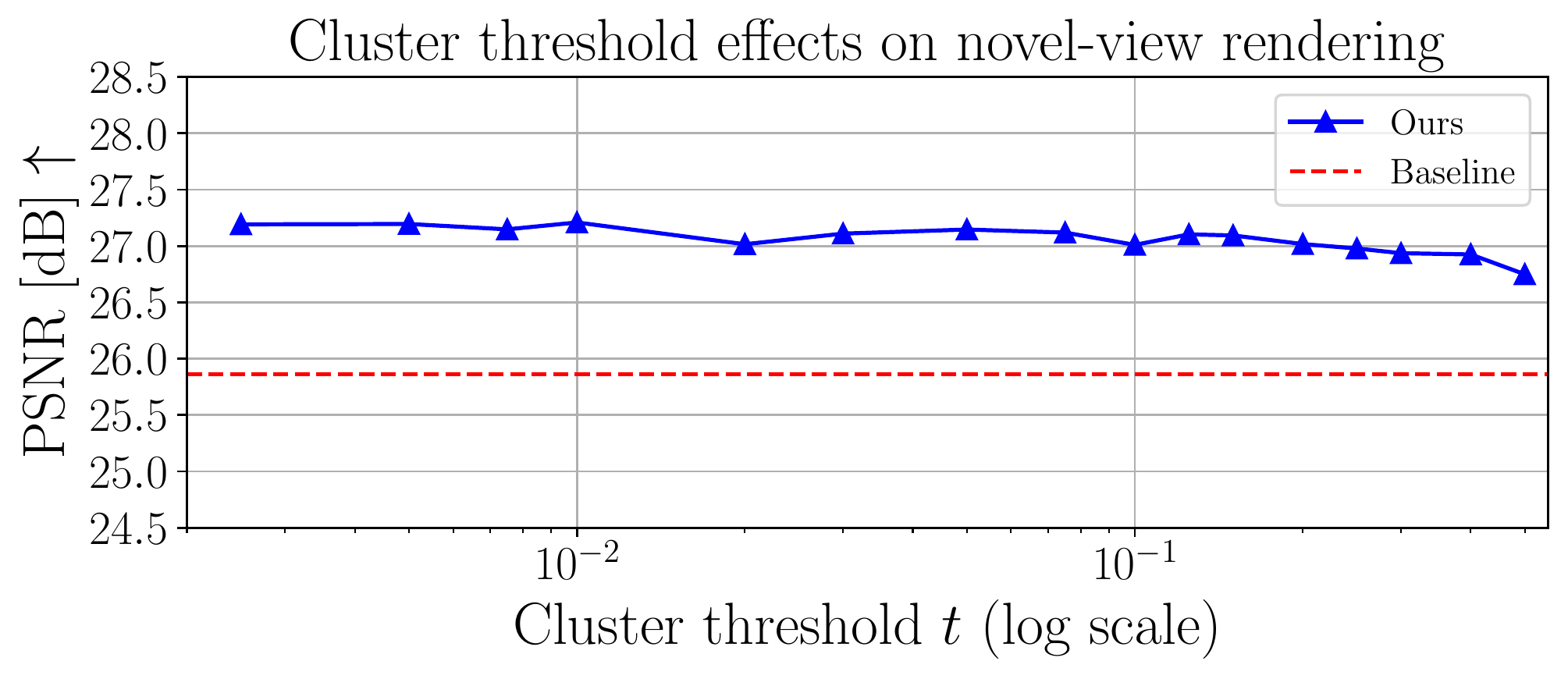}
        \caption{Cluster threshold choice effects on novel-view rendering.}
        \label{fig:supp:cluster_thresh_psnr}
    \end{subfigure}
    \begin{subfigure}[b]{\columnwidth}
        \centering
        \includegraphics[width=\columnwidth]{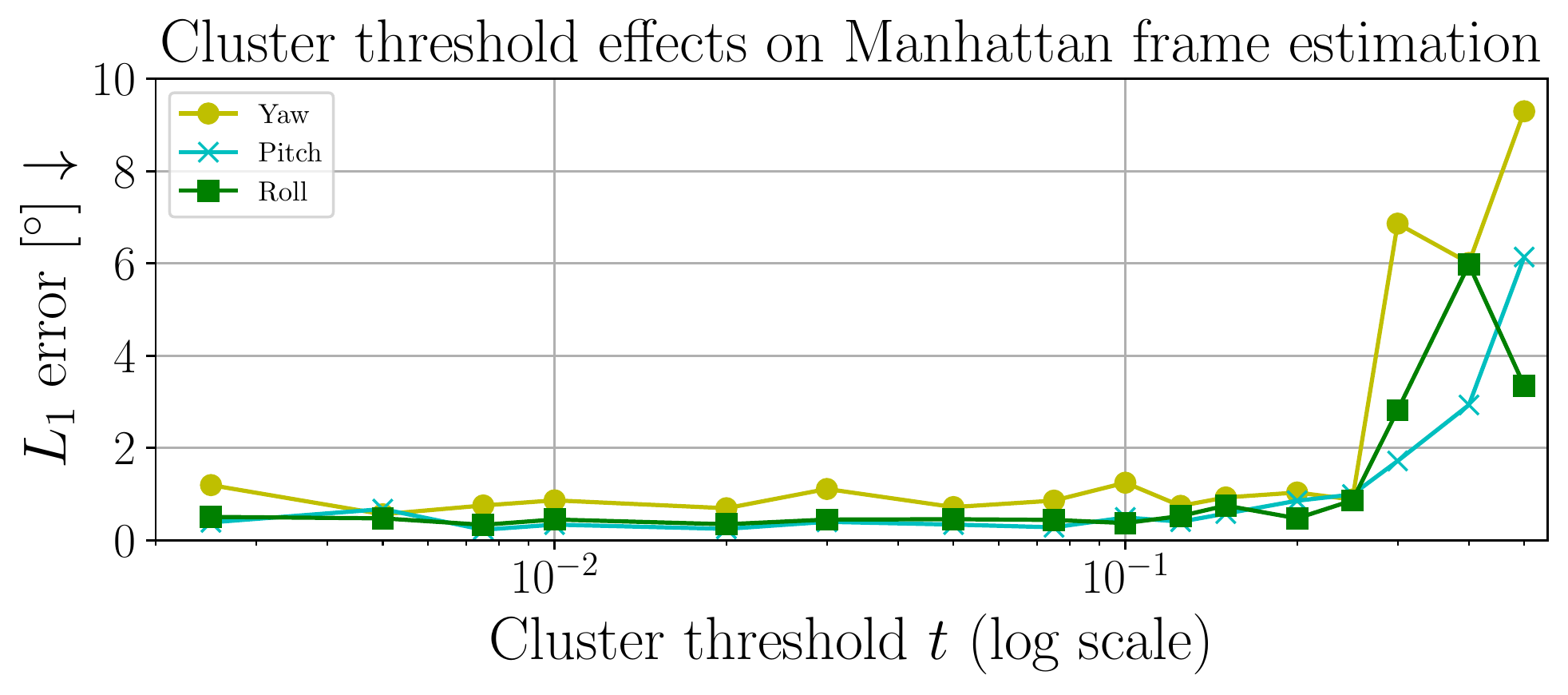}
        \caption{Cluster threshold choice effects on estimating the MF.}
        \label{fig:supp:cluster_thresh_angles}
    \end{subfigure}
    \caption{{\textbf{Cluster threshold $t$ choice effects on Hypersim-A.} Our method is not very sensitive to threshold values below $0.1$.}}
    \label{fig:supp:cluster_thresh}
\end{figure}
\noindent \textbf{Cluster threshold:} We analyze the sensitivity of our proposed method to the threshold parameter $t$, used to merge selected orthogonal clusters with their opposites and their nearby clusters. In Figure~\ref{fig:supp:cluster_thresh}, we see that our algorithm is not very sensitive to different threshold $t$ values below $0.1$.

\begin{table}[b!]
\centering
\tableFont
\captionsetup{font=small}
\caption{\textbf{Ray sampling strategy effects Hypersim-A.} There is no significant difference for the InstantNGP baseline between sampling random triplet triangles or random rays during training.}
\begin{tabular}{|c|c|c|c|c|c|}
\hline
 & PSNR$\uparrow$  & D$\downarrow$-MAE & Norm. $\degree\downarrow$ & Rot. $\degree\downarrow$ \\ \hline \hline
\makecell{Random rays \\(same image)}
& 25.72 & 0.072 & 60.11 & 12.36 \\  \hline
\makecell{Random rays \\ (random images)}
& 25.86 & 0.064 & 57.12 & 10.63 \\  \hline
\makecell{Random triangles \\ (random images)}
& 26.16 & 0.061 & 56.72 & 10.25 \\  \hline

\end{tabular}
\label{table:supp:ray_sampling}
\end{table}
\noindent \textbf{Ray sampling:} As previously mentioned, when using methods which rely on explicit surface normals, we randomly sample pixel triplet triangles from a set of all possible triangles in all available images. However, when using the InstantNGP baseline, we randomly sample rays from the set of all available rays in all available images. The number of selected rays is always the same in total as the specified batch size, for all methods. In Table~\ref{table:supp:ray_sampling}, we see that there is no significant difference if we sample triplet triangles during the baseline training instead of just randomly sampling rays. Therefore the improvements in our approach come from the proposed algorithm, and not the slightly different way of batch sampling.

\begin{figure}[t!]
    \centering
    \captionsetup{font=small}
    \begin{subfigure}[b]{\columnwidth}
        \centering
        \includegraphics[width=\columnwidth]{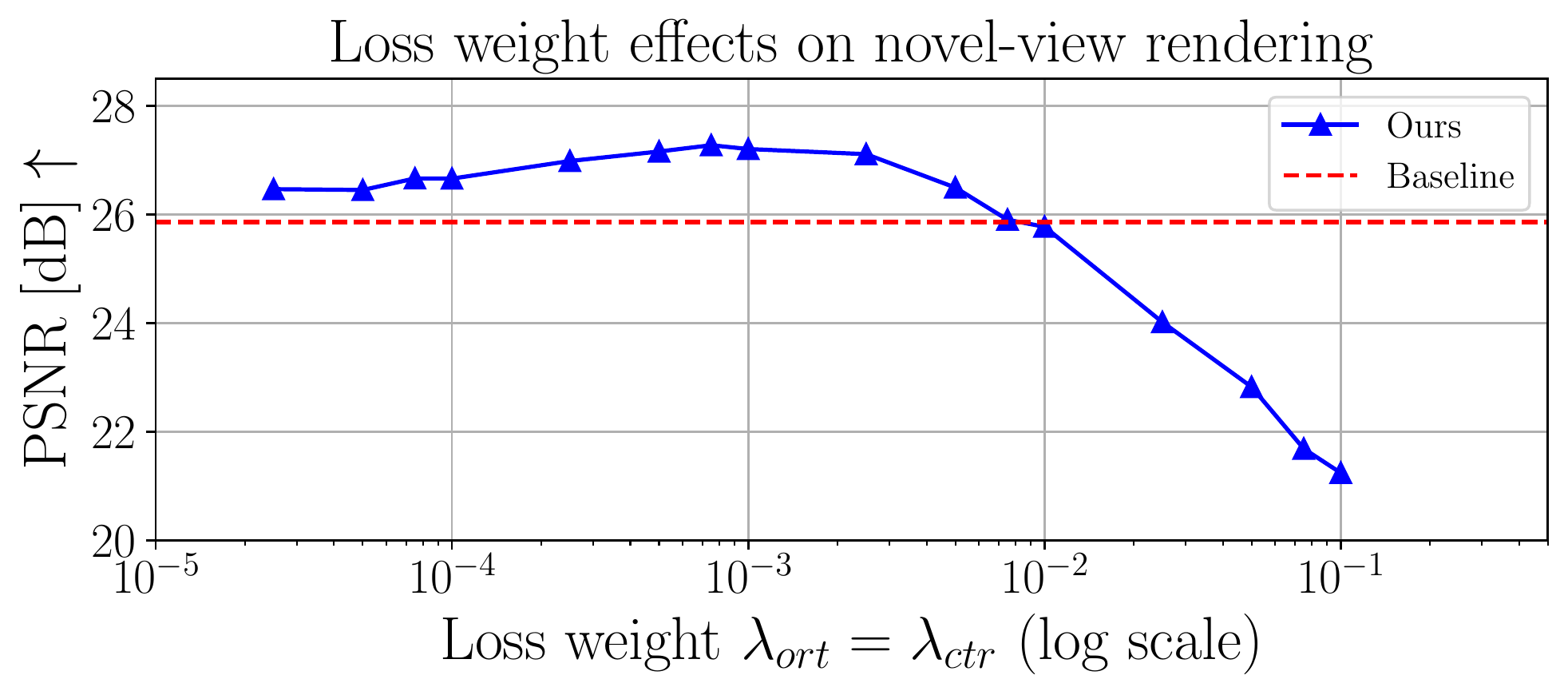}
        \caption{Loss weight effects on novel-view rendering.}
        \label{fig:supp:loss_w_psnr}
    \end{subfigure}
    \begin{subfigure}[b]{\columnwidth}
        \centering
        \includegraphics[width=\columnwidth]{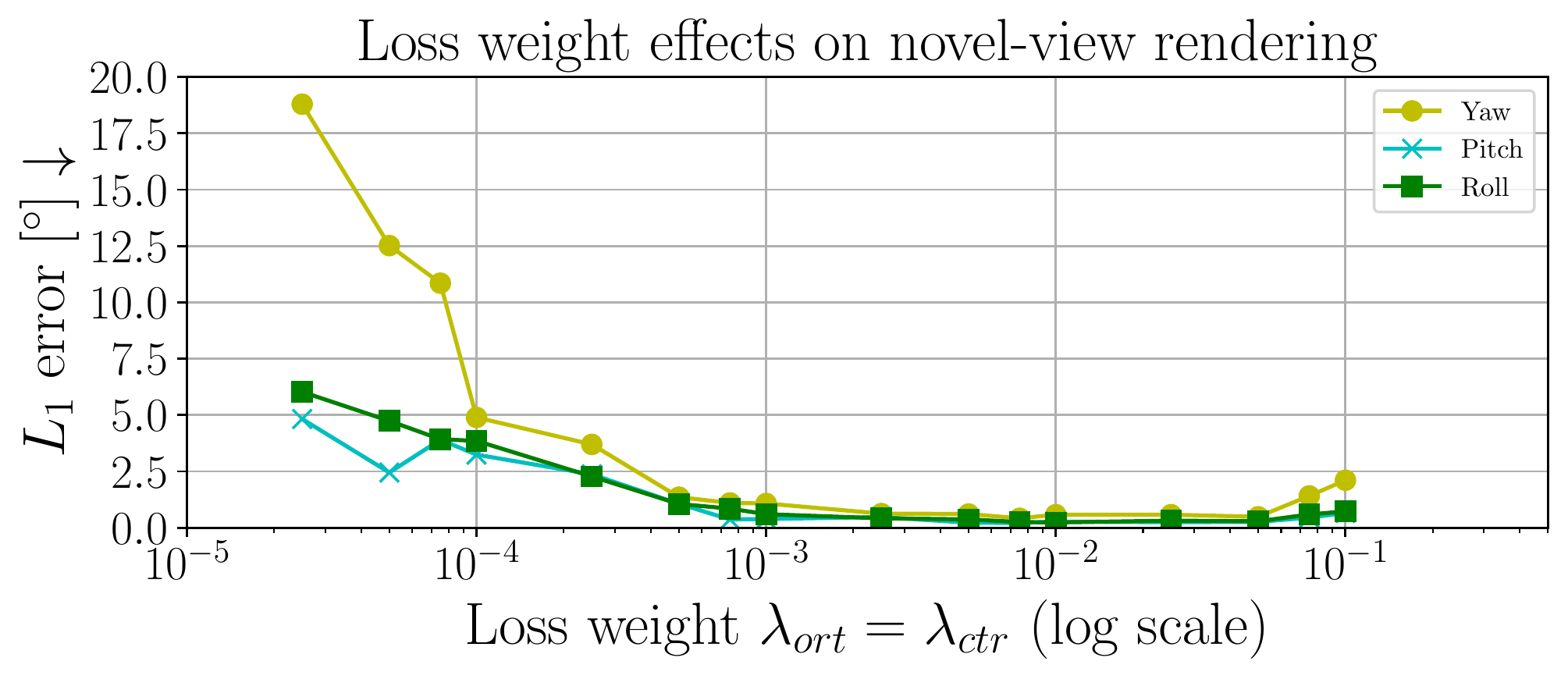}
        \caption{Loss weight effects on estimating the MF.}
        \label{fig:supp:loss_w_angles}
    \end{subfigure}
    \caption{{\textbf{Loss weight $\lambda_{ort}=\lambda_{ctr}$ choice effects on Hypersim-A.} Very low weights lead to bad MF estimation, whereas very high weights lead to bad novel-view rendering.}}
    \label{fig:supp:loss_w}
\end{figure}
\noindent \textbf{Sensitivity in  hyperparameters selection:} We further analyse the sensitivity of the loss weights $\lambda_{ort}=\lambda_{ctr}$ multiplying our proposed loss terms, to the overall performance. The quantitative results are presented in Figure~\ref{fig:supp:loss_w}. Very low $\lambda$ values lead to a bad MF estimation, whereas very high $\lambda$ values lead to bad novel-view rendering. A good trade-off is achieved around the chosen value of $2 \cdot 10^{-3}$.
In addition, visual results are presented in Figure~\ref{fig:supp:vis_loss_w}, where we can observe that very low $\lambda$ produces noisy explicit surface normals, whereas very high $\lambda$ makes all details appear ``blocky" in normals. Moreover, this analysis has been performed statistically, on 20 scenes from the Hypersim-A split using the same hyperparameters. We observe that the scene-specific tuning of these hyperparameters could further improve the performance. These scene-specific hyperparameters differ only slightly from the ones selected for the whole set of scenes. We however, do not perform scene-specific tuning, demonstrating the generaliziblity of our method  across the diverse scenes of Hypersim.

\noindent\textbf{Runtime: } We further analyze the run time of our proposed method. We use Python and the PyTorch Deep Learning library. The InstantNGP baseline implementation is adapted from~\cite{ngp_pl_git}, where some functionalities (e.g. volume rendering) are efficiently implemented directly in CUDA code. In addition to the baseline, our proposed approach computes explicit surface normals for every batch, followed by k-means clustering (implemented using the FIASS library~\cite{johnson2019billion}), and finally followed by computing two additional loss terms $\mathcal{L}_{ctr}$ and $\mathcal{L}_{ort}$.  The baseline needs $22.77 \pm 2.07$ minutes on average to train on Hypersim-A, whereas our method needs $29.06 \pm 1.64$ minutes. The inference time is the same for both methods, and it takes about half a minute on average to render the test set. The experiments were performed on a single NVIDIA GeForce RTX 2080 Ti GPU, with 11 Gb of RAM memory.

\section{Limitations}
\label{sec:supp:limitations}
\begin{figure}[hb!]
\centering
\addtolength{\tabcolsep}{-4.5pt}
\begin{tabular} {c||cc}
 & Ours & GT\\ \hline \hline

 \multirow{1}{*}[20pt]{\rotatebox{90}{RGB}} &
 \includegraphics[width=0.44\columnwidth]{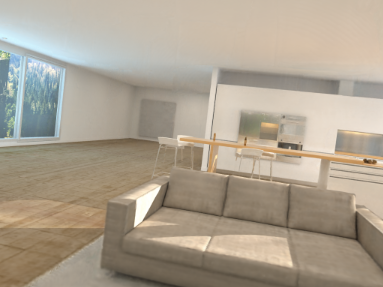}&
 \includegraphics[width=0.44\columnwidth]{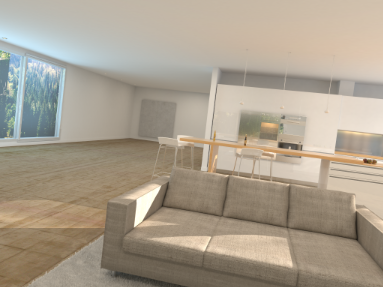} \\
 
 \multirow{1}{*}[40pt]{\rotatebox{90}{Normals}} &
 \includegraphics[width=0.44\columnwidth]{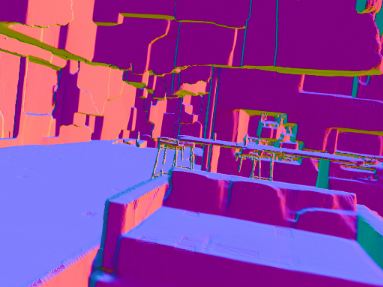}&
 \includegraphics[width=0.44\columnwidth]{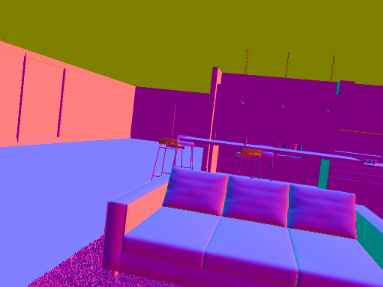} \\ \hline \hline 

 \multirow{1}{*}[20pt]{\rotatebox{90}{RGB}} &
 \includegraphics[width=0.44\columnwidth]{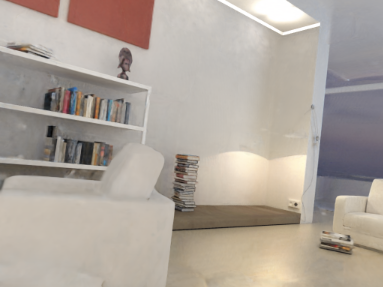}&
 \includegraphics[width=0.44\columnwidth]{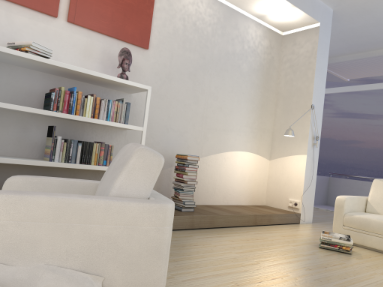} \\
 
 \multirow{1}{*}[40pt]{\rotatebox{90}{Normals}} &
 \includegraphics[width=0.44\columnwidth]{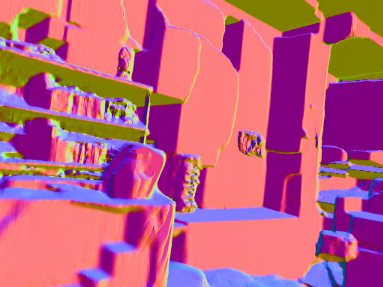}&
 \includegraphics[width=0.44\columnwidth]{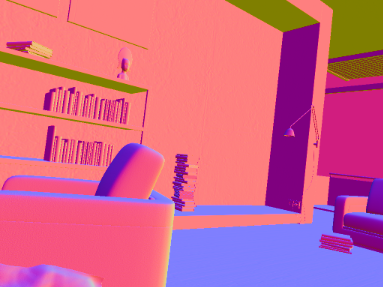} \\ 
 
\end{tabular}
	\caption{\textbf{Common failure cases.} In the first example, we see our method merging explicitly computed surface normals of one Manhattan direction (vertical ceiling surface) with one of the other two directions (horizontal wall surface). In the second example, we see severe ``blocky" artifacts in the normals. Nevertheless, in both cases, there is not much trouble, nor big artifacts, when performing novel-view rendering.}
	\label{fig:supp:vis_fails}
\end{figure}
\begin{figure}[t!]
\centering
\addtolength{\tabcolsep}{-4.5pt}
\begin{tabular} {c||cc}
 & Normals & RGB\\ \hline \hline
 
 \multirow{1}{*}[40pt]{\rotatebox{90}{$\lambda=0$}} &
 \includegraphics[width=0.44\linewidth]{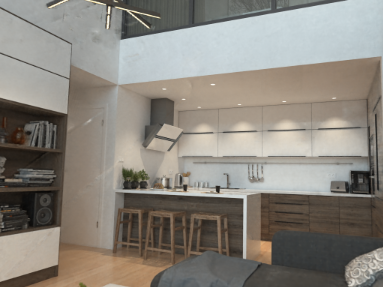}&
 \includegraphics[width=0.44\linewidth]{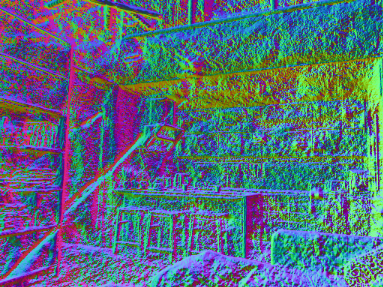} \\
 
 \multirow{1}{*}[60pt]{\rotatebox{90}{$\lambda=5\cdot10^{-5}$}} &
 \includegraphics[width=0.44\linewidth]{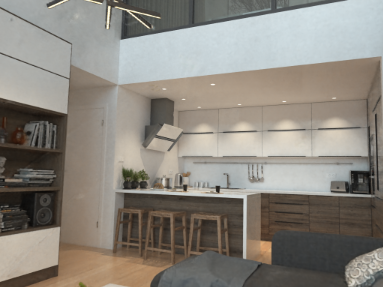}&
 \includegraphics[width=0.44\linewidth]{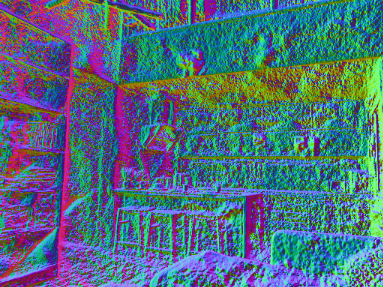} \\
 
 \multirow{1}{*}[60pt]{\rotatebox{90}{$\lambda=5\cdot10^{-4}$}} &
 \includegraphics[width=0.44\linewidth]{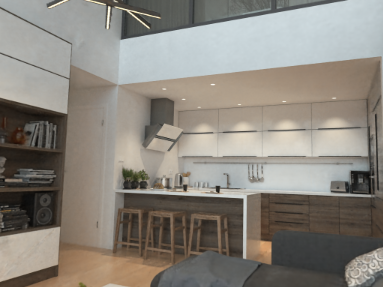}&
 \includegraphics[width=0.44\linewidth]{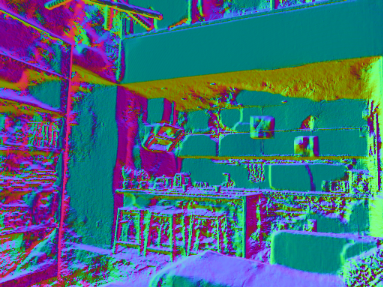} \\

 \multirow{1}{*}[60pt]{\rotatebox{90}{$\lambda=2\cdot10^{-3}$}} &
 \includegraphics[width=0.44\linewidth]{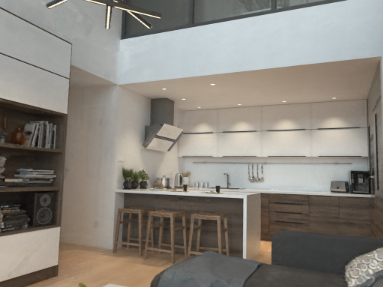}&
 \includegraphics[width=0.44\linewidth]{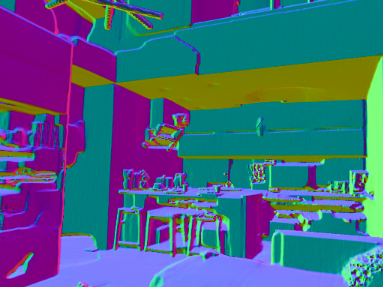} \\
 
 \multirow{1}{*}[50pt]{\rotatebox{90}{$\lambda=10^{-2}$}} &
 \includegraphics[width=0.44\linewidth]{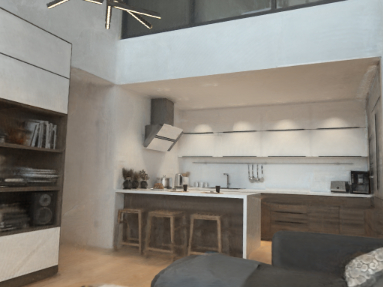}&
 \includegraphics[width=0.44\linewidth]{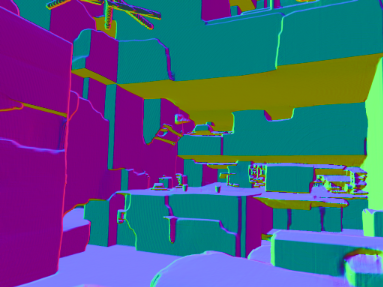} \\
 
 \multirow{1}{*}[50pt]{\rotatebox{90}{$\lambda=10^{-1}$}} &
 \includegraphics[width=0.44\linewidth]{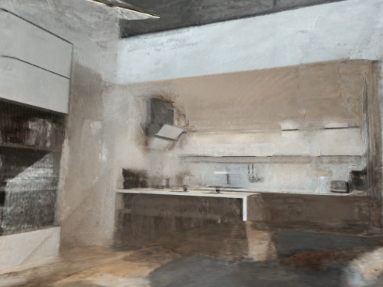}&
 \includegraphics[width=0.44\linewidth]{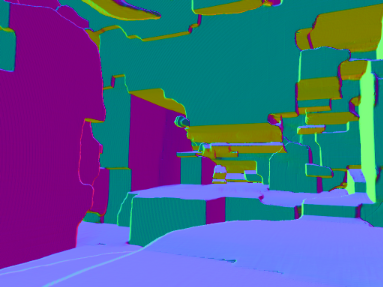} \\
 
 \multirow{1}{*}[40pt]{\rotatebox{90}{GT}} &
 \includegraphics[width=0.44\linewidth]{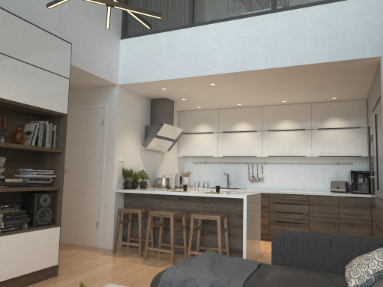}&
 \includegraphics[width=0.44\linewidth]{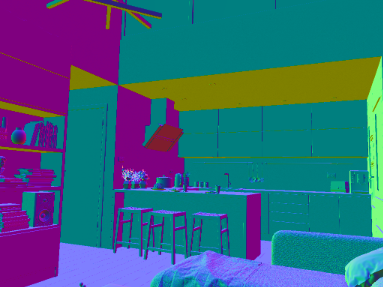} \\
 
\end{tabular}
	\caption{\textbf{Loss weight $\lambda_{ort}=\lambda_{ctr}$ choice effects on Hypersim-A.} Very low $\lambda$ values lead to very noisy explicit surface normals, whereas very high $\lambda$ values lead to "blocky" artifacts in normals and bad novel-view rendering.}
	\label{fig:supp:vis_loss_w}
\end{figure}

\noindent\textbf{Missing details in surface normals:}
While inspecting qualitative results of novel-view rendering, we observed a few limitations and failure cases that occurred occasionally. Sometimes our method merges one Manhattan direction with one of the other two directions, e.g. producing the same surface normals for a roof (horizontal surface) and one of the walls (vertical surface). This can be seen in the first example in Figure~\ref{fig:supp:vis_fails}. Another phenomenon is having severe ``blocky" artifacts or missing details in the computed explicit normals, e.g. as in the second example in Figure~\ref{fig:supp:vis_fails}. 
Nevertheless, this usually offers better novel-view RGB rendering, compared to not imposing any structure priors.
This behavior is also related to the choice of loss weights $\lambda_{ort}=\lambda_{ctr}$, discussed in Section~\ref{sec:supp:further_analysis} of this supplementary material, as well as in Figure~\ref{fig:supp:vis_loss_w}.

\noindent\textbf{Easy scenes:}
Our method is not beneficial for very easy scenes. This is shown in
\ifx\FileIsMerged\undefined
Figure~5 of the main paper.
\else
Figure~\ref{fig:improvement_wrt_difficulty} of the main paper.
\fi
When it comes to easy scenes, the 3D structure of these scenes can be learned without much problem, without imposing any structure priors.

\section{More Qualitative Results}
\label{sec:supp:qualitative_results}
\begin{figure*}[t!]
\centering
\addtolength{\tabcolsep}{-4.5pt}
\begin{tabular} {c||cccc}
 & InstaNGP & ManDF & Ours & GT\\ \hline \hline

 \multirow{1}{*}[20pt]{\rotatebox{90}{RGB}} &
 \includegraphics[width=0.17\linewidth]{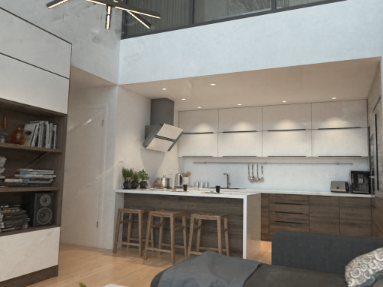}&
 \includegraphics[width=0.17\linewidth]{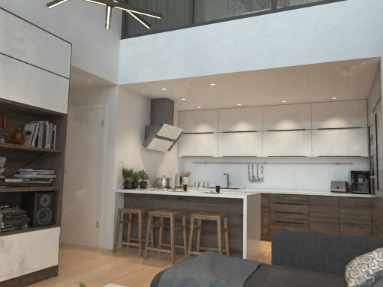}&
 \includegraphics[width=0.17\linewidth]{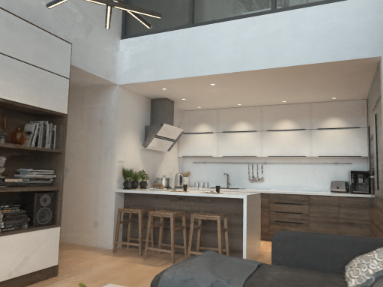}&
 \includegraphics[width=0.17\linewidth]{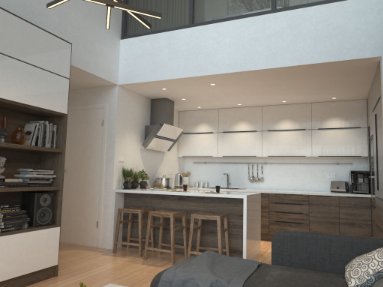} \\
 
 \multirow{1}{*}[25pt]{\rotatebox{90}{Depth}} &
 \includegraphics[width=0.17\linewidth]{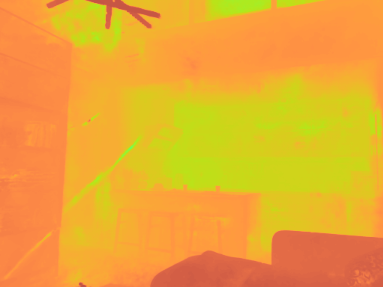}&
 \includegraphics[width=0.17\linewidth]{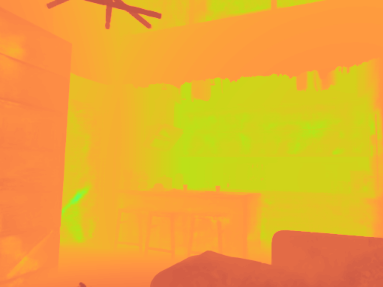}&
 \includegraphics[width=0.17\linewidth]{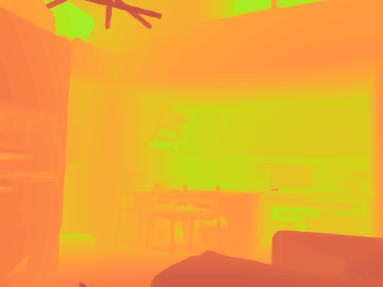}&
 \includegraphics[width=0.17\linewidth]{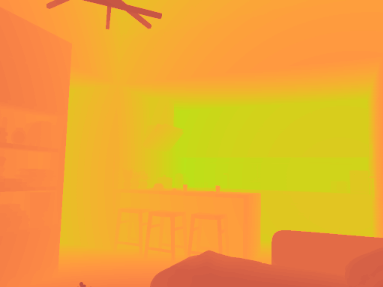} \\
 
 \multirow{1}{*}[30pt]{\rotatebox{90}{Normals}} &
 \includegraphics[width=0.17\linewidth]{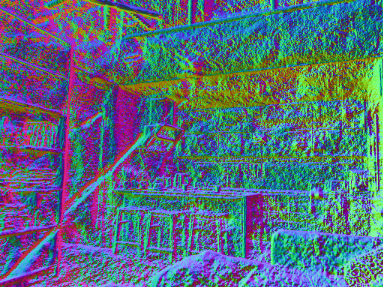}&
 \includegraphics[width=0.17\linewidth]{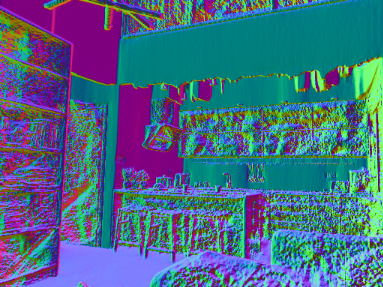}&
 \includegraphics[width=0.17\linewidth]{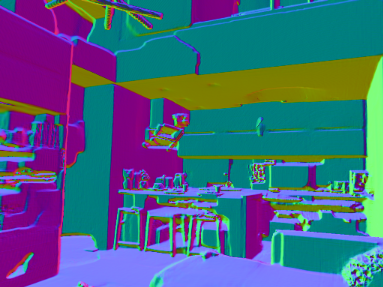}&
 \includegraphics[width=0.17\linewidth]{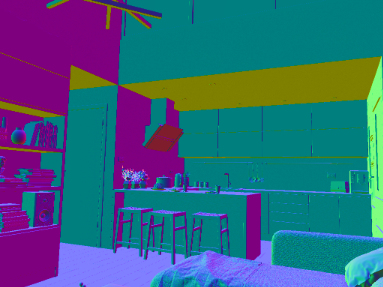} \\ \hline \hline
 
 \multirow{1}{*}[20pt]{\rotatebox{90}{RGB}} &
 \includegraphics[width=0.17\linewidth]{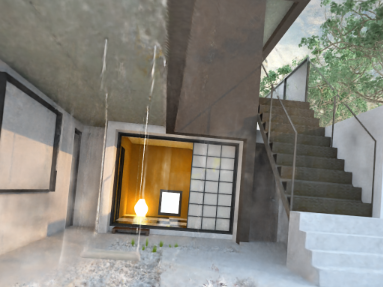}&
 \includegraphics[width=0.17\linewidth]{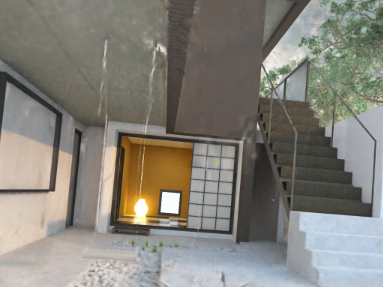}&
 \includegraphics[width=0.17\linewidth]{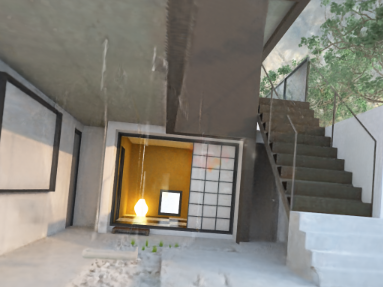}&
 \includegraphics[width=0.17\linewidth]{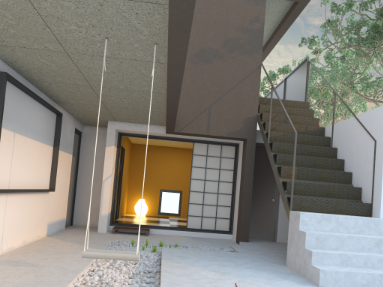} \\
 
 \multirow{1}{*}[30pt]{\rotatebox{90}{Normals}} &
 \includegraphics[width=0.17\linewidth]{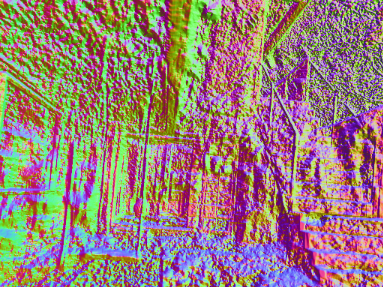}&
 \includegraphics[width=0.17\linewidth]{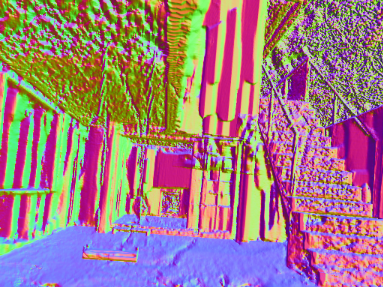}&
 \includegraphics[width=0.17\linewidth]{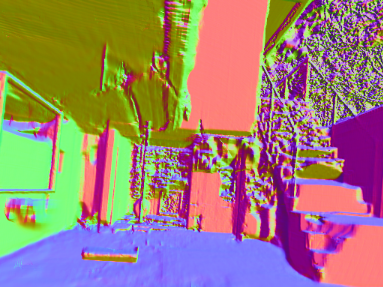}&
 \includegraphics[width=0.17\linewidth]{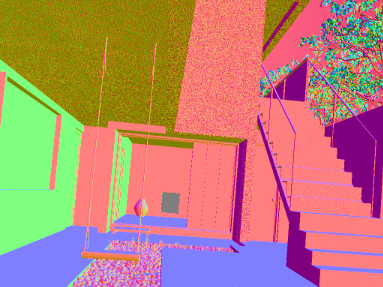} \\ \hline \hline
 
  \multirow{1}{*}[20pt]{\rotatebox{90}{RGB}} &
 \includegraphics[width=0.17\linewidth]{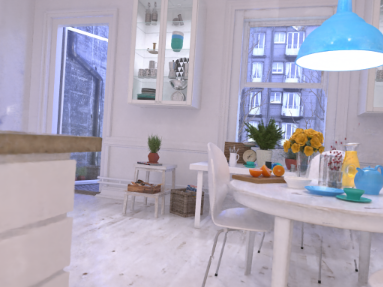}&
 \includegraphics[width=0.17\linewidth]{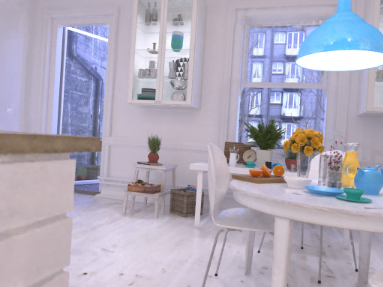}&
 \includegraphics[width=0.17\linewidth]{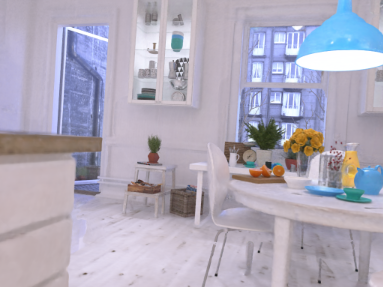}&
 \includegraphics[width=0.17\linewidth]{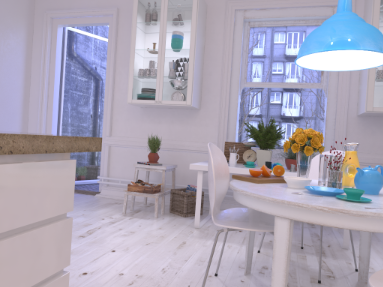} \\
 
 \multirow{1}{*}[30pt]{\rotatebox{90}{Normals}} &
 \includegraphics[width=0.17\linewidth]{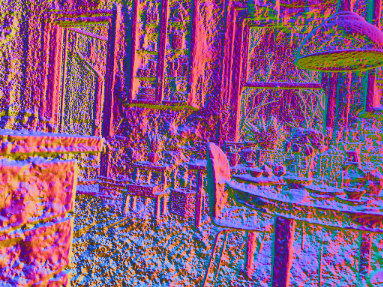}&
 \includegraphics[width=0.17\linewidth]{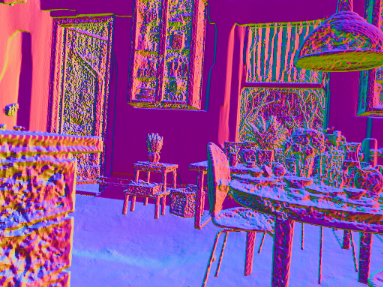}&
 \includegraphics[width=0.17\linewidth]{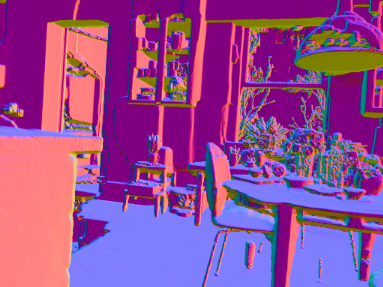}&
 \includegraphics[width=0.17\linewidth]{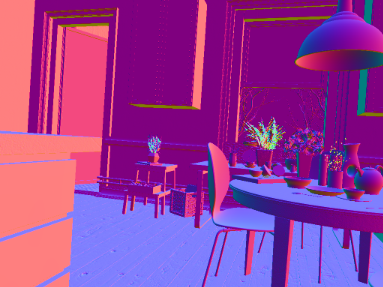} \\ \hline \hline
 
 \multirow{1}{*}[20pt]{\rotatebox{90}{RGB}} &
 \includegraphics[width=0.17\linewidth]{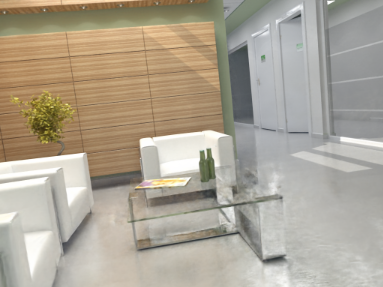}&
 \includegraphics[width=0.17\linewidth]{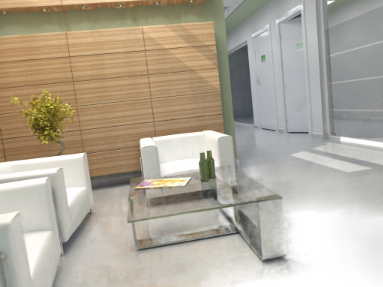}&
 \includegraphics[width=0.17\linewidth]{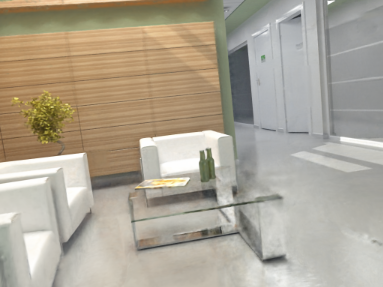}&
 \includegraphics[width=0.17\linewidth]{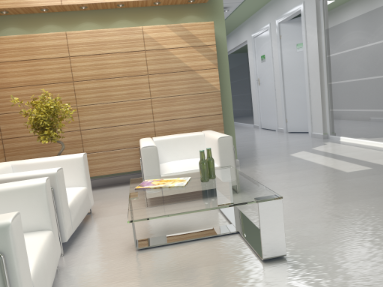} \\
 
 \multirow{1}{*}[30pt]{\rotatebox{90}{Normals}} &
 \includegraphics[width=0.17\linewidth]{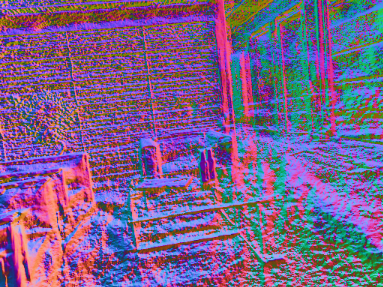}&
 \includegraphics[width=0.17\linewidth]{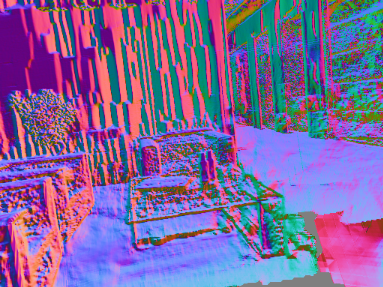}&
 \includegraphics[width=0.17\linewidth]{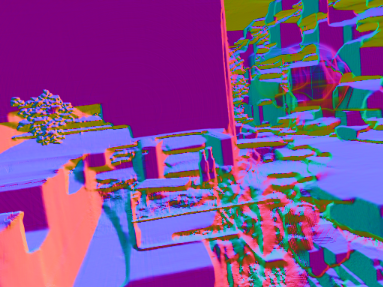}&
 \includegraphics[width=0.17\linewidth]{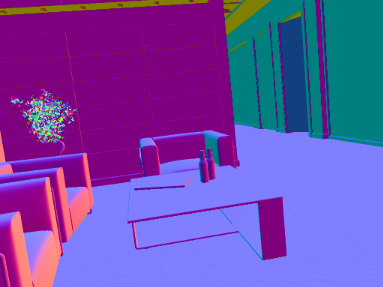} \\
 
\end{tabular}
	\caption{\textbf{Qualitative results on Hypersim-A.}  Our method leverages many Manhattan objects and surfaces in the scene, which improves the implicit geometrical representation compared to the baseline. Unlike ManhattanDF, our method relies on many Manhattan cues other than the walls and floors.}
	\label{fig:supp:vis_hypersim}
\end{figure*}

\begin{figure*}[t!]
\centering
\addtolength{\tabcolsep}{-4.5pt}
\begin{tabular} {c||cccc}
 & InstaNGP & ManDF & Ours & GT\\ \hline \hline

 \multirow{1}{*}[20pt]{\rotatebox{90}{RGB}} &
 \includegraphics[width=0.17\linewidth]{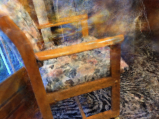}&
 \includegraphics[width=0.17\linewidth]{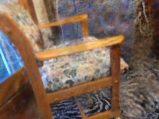}&
 \includegraphics[width=0.17\linewidth]{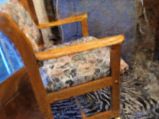}&
 \includegraphics[width=0.17\linewidth]{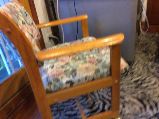} \\
 
 \multirow{1}{*}[25pt]{\rotatebox{90}{Depth}} &
 \includegraphics[width=0.17\linewidth]{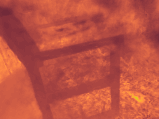}&
 \includegraphics[width=0.17\linewidth]{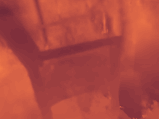}&
 \includegraphics[width=0.17\linewidth]{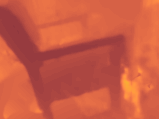}&
 \includegraphics[width=0.17\linewidth]{images/vis_results/not_available.png} \\
 
 \multirow{1}{*}[30pt]{\rotatebox{90}{Normals}} &
 \includegraphics[width=0.17\linewidth]{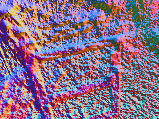}&
 \includegraphics[width=0.17\linewidth]{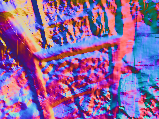}&
 \includegraphics[width=0.17\linewidth]{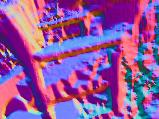}&
 \includegraphics[width=0.17\linewidth]{images/vis_results/not_available.png} \\ \hline \hline
 
 \multirow{1}{*}[20pt]{\rotatebox{90}{RGB}} &
 \includegraphics[width=0.17\linewidth]{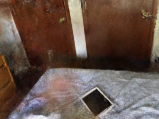}&
 \includegraphics[width=0.17\linewidth]{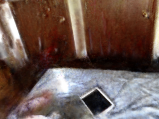}&
 \includegraphics[width=0.17\linewidth]{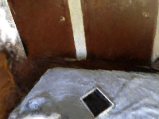}&
 \includegraphics[width=0.17\linewidth]{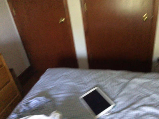} \\
 
 \multirow{1}{*}[30pt]{\rotatebox{90}{Normals}} &
 \includegraphics[width=0.17\linewidth]{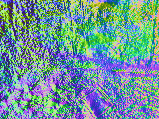}&
 \includegraphics[width=0.17\linewidth]{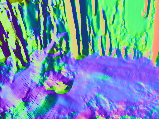}&
 \includegraphics[width=0.17\linewidth]{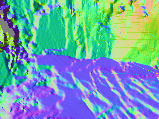}&
 \includegraphics[width=0.17\linewidth]{images/vis_results/not_available.png} \\ \hline \hline
 
 \multirow{1}{*}[20pt]{\rotatebox{90}{RGB}} &
 \includegraphics[width=0.17\linewidth]{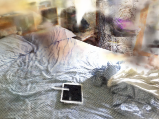}&
 \includegraphics[width=0.17\linewidth]{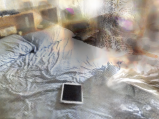}&
 \includegraphics[width=0.17\linewidth]{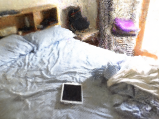}&
 \includegraphics[width=0.17\linewidth]{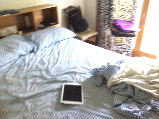} \\
 
 \multirow{1}{*}[30pt]{\rotatebox{90}{Normals}} &
 \includegraphics[width=0.17\linewidth]{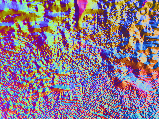}&
 \includegraphics[width=0.17\linewidth]{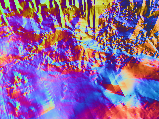}&
 \includegraphics[width=0.17\linewidth]{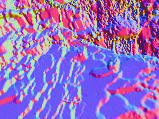}&
 \includegraphics[width=0.17\linewidth]{images/vis_results/not_available.png} \\ \hline \hline
 
 \multirow{1}{*}[20pt]{\rotatebox{90}{RGB}} &
 \includegraphics[width=0.17\linewidth]{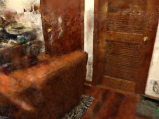}&
 \includegraphics[width=0.17\linewidth]{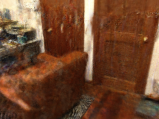}&
 \includegraphics[width=0.17\linewidth]{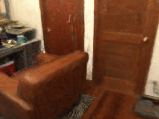}&
 \includegraphics[width=0.17\linewidth]{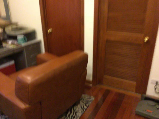} \\
 
 \multirow{1}{*}[30pt]{\rotatebox{90}{Normals}} &
 \includegraphics[width=0.17\linewidth]{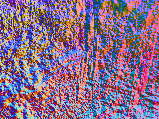}&
 \includegraphics[width=0.17\linewidth]{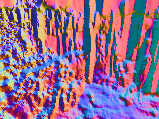}&
 \includegraphics[width=0.17\linewidth]{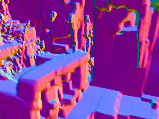}&
 \includegraphics[width=0.17\linewidth]{images/vis_results/not_available.png} \\
 
\end{tabular}
\caption{\textbf{Qualitative results on ScanNet.} Our method leverages many Manhattan objects and surfaces in the scene, which improves the implicit geometrical representation compared to the baseline. Unlike ManhattanDF, our method relies on many Manhattan cues other than the walls and floors.}
\label{fig:supp:scannet}
\end{figure*}

\begin{figure*}[t!]
\centering
\addtolength{\tabcolsep}{-4.5pt}
\begin{tabular} {c||cccc}
 & InstaNGP & ManDF & Ours & GT\\ \hline \hline

 \multirow{1}{*}[20pt]{\rotatebox{90}{RGB}} &
 \includegraphics[width=0.17\linewidth]{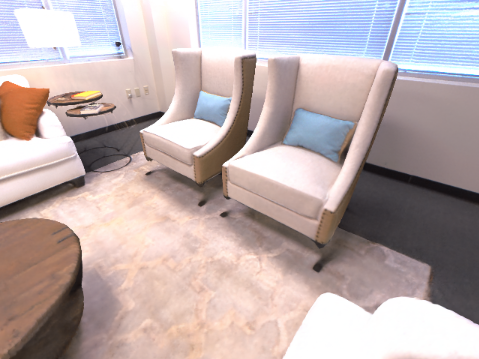}&
 \includegraphics[width=0.17\linewidth]{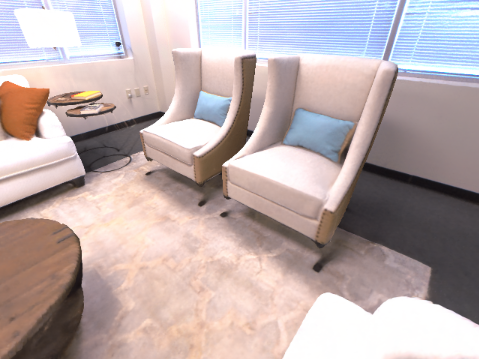}&
 \includegraphics[width=0.17\linewidth]{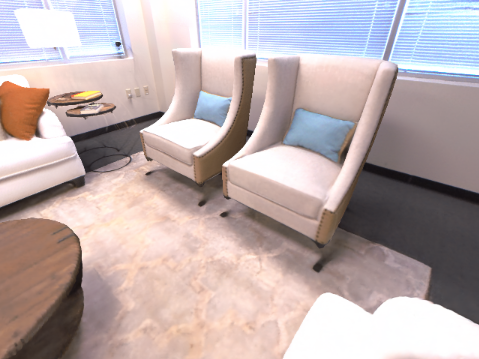}&
 \includegraphics[width=0.17\linewidth]{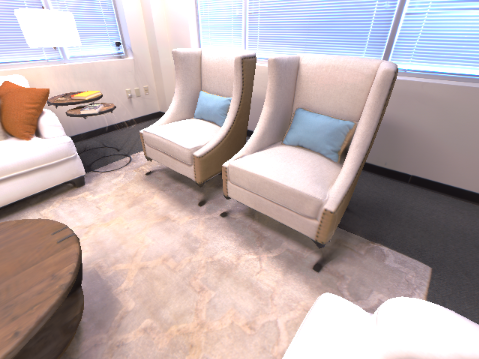} \\
 
 \multirow{1}{*}[25pt]{\rotatebox{90}{Depth}} &
 \includegraphics[width=0.17\linewidth]{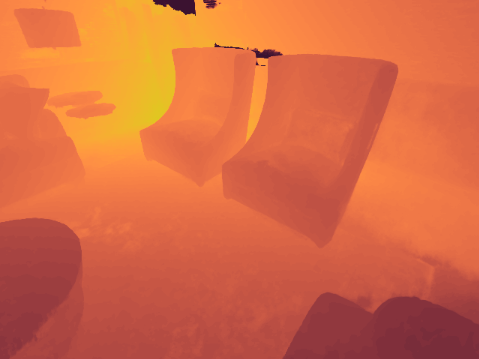}&
 \includegraphics[width=0.17\linewidth]{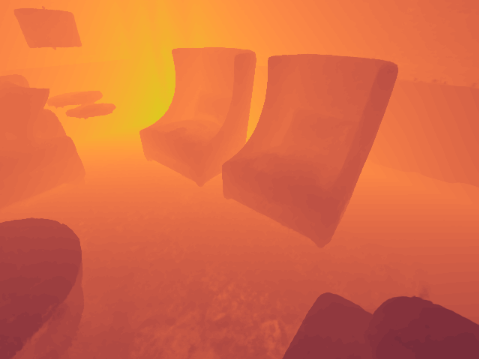}&
 \includegraphics[width=0.17\linewidth]{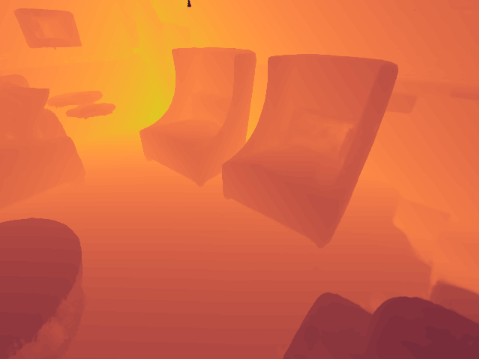}&
 \includegraphics[width=0.17\linewidth]{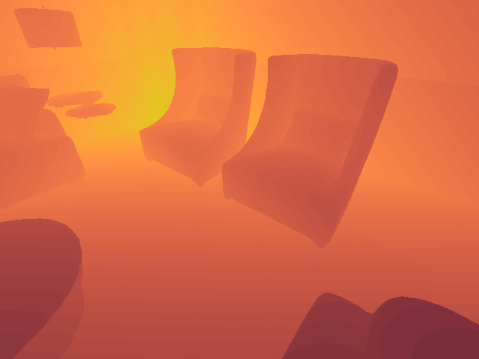} \\
 
 \multirow{1}{*}[30pt]{\rotatebox{90}{Normals}} &
 \includegraphics[width=0.17\linewidth]{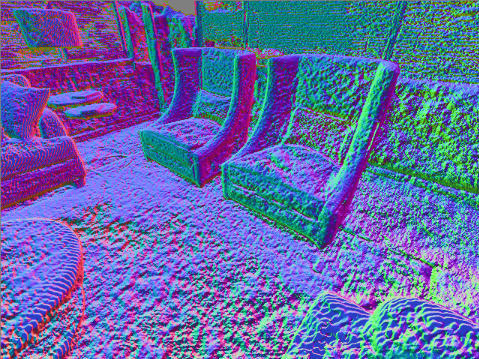}&
 \includegraphics[width=0.17\linewidth]{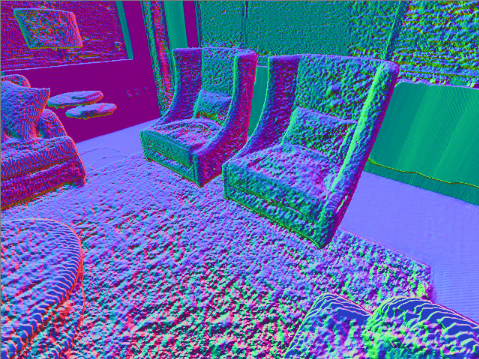}&
 \includegraphics[width=0.17\linewidth]{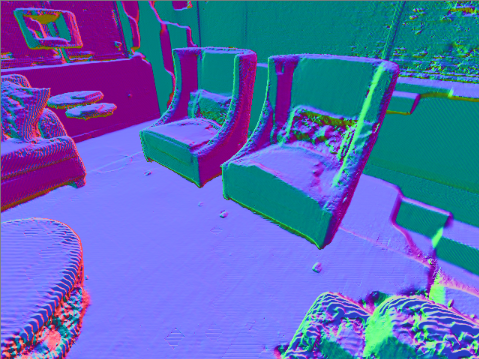}&
 \includegraphics[width=0.17\linewidth]{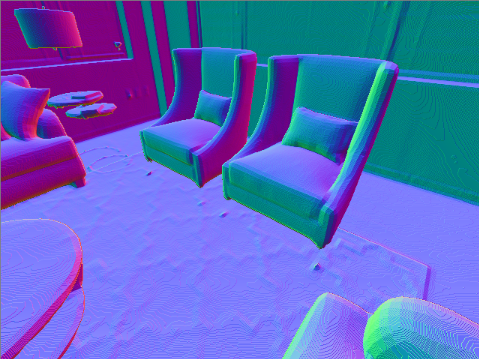} \\ \hline \hline
 
 \multirow{1}{*}[20pt]{\rotatebox{90}{RGB}} &
 \includegraphics[width=0.17\linewidth]{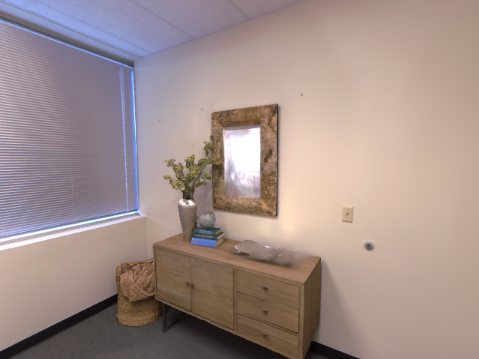}&
 \includegraphics[width=0.17\linewidth]{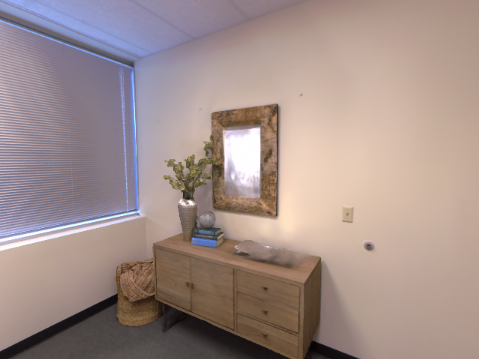}&
 \includegraphics[width=0.17\linewidth]{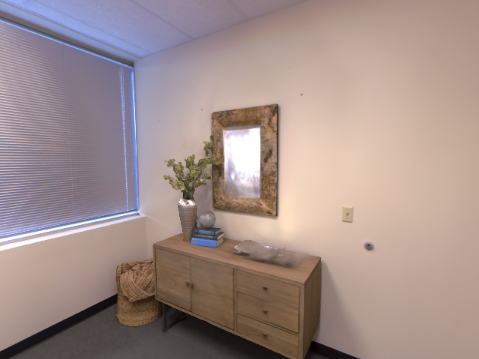}&
 \includegraphics[width=0.17\linewidth]{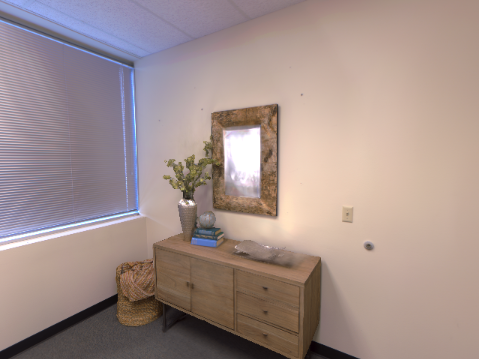} \\
 
 \multirow{1}{*}[30pt]{\rotatebox{90}{Normals}} &
 \includegraphics[width=0.17\linewidth]{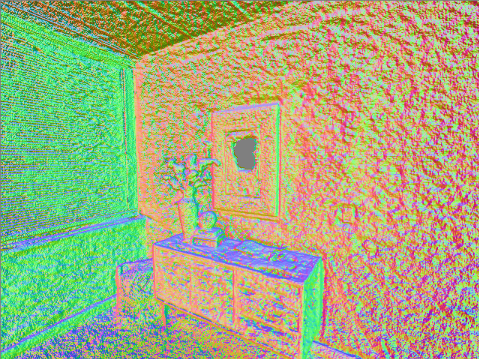}&
 \includegraphics[width=0.17\linewidth]{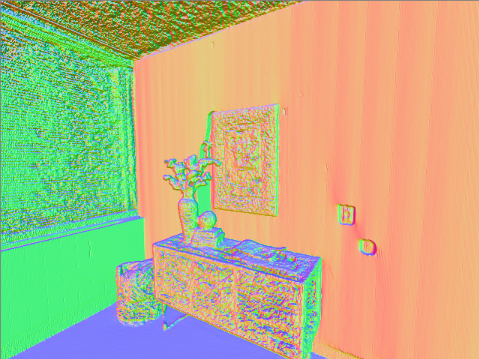}&
 \includegraphics[width=0.17\linewidth]{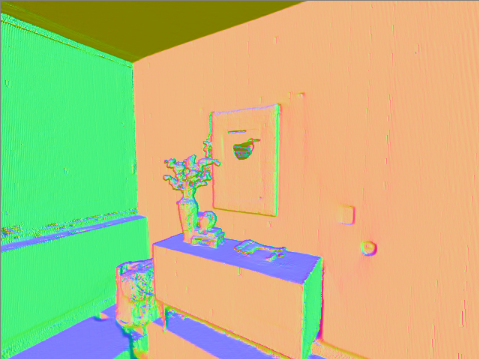}&
 \includegraphics[width=0.17\linewidth]{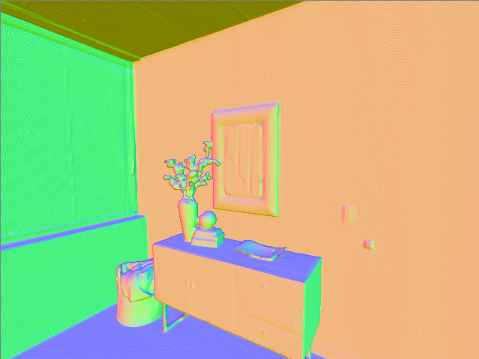} \\ \hline \hline
 
 \multirow{1}{*}[20pt]{\rotatebox{90}{RGB}} &
 \includegraphics[width=0.17\linewidth]{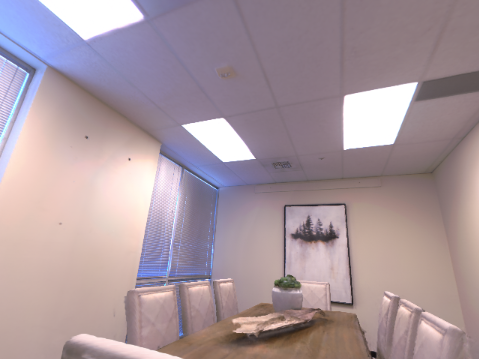}&
 \includegraphics[width=0.17\linewidth]{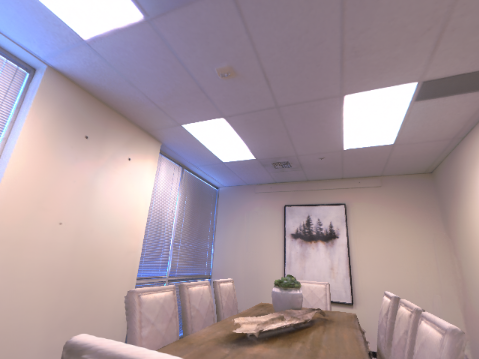}&
 \includegraphics[width=0.17\linewidth]{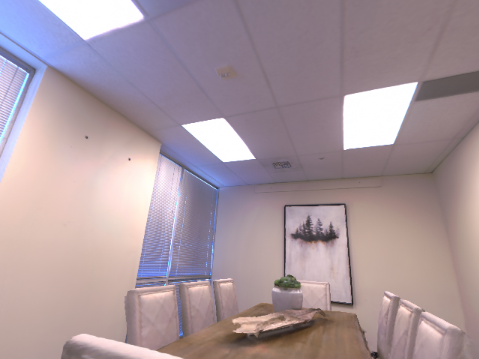}&
 \includegraphics[width=0.17\linewidth]{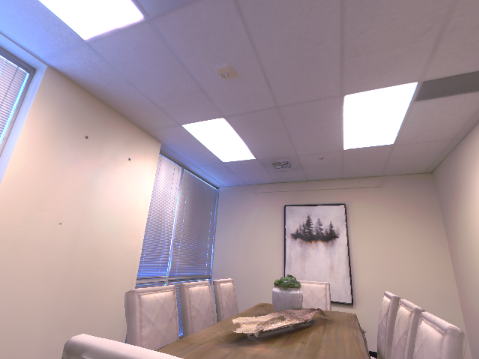} \\
 
 \multirow{1}{*}[30pt]{\rotatebox{90}{Normals}} &
 \includegraphics[width=0.17\linewidth]{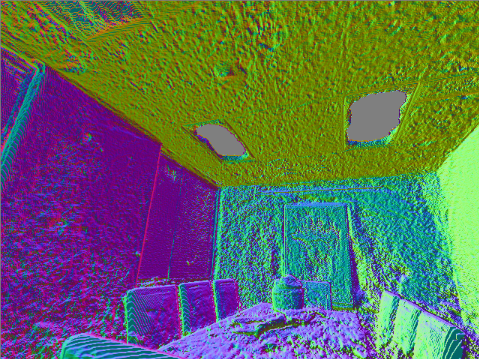}&
 \includegraphics[width=0.17\linewidth]{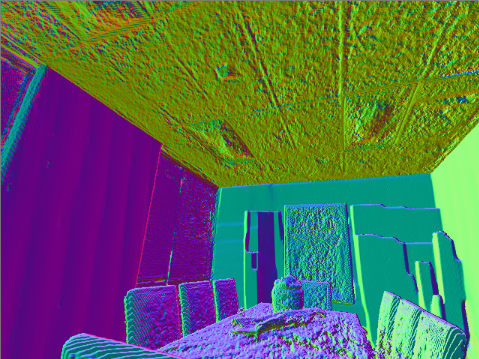}&
 \includegraphics[width=0.17\linewidth]{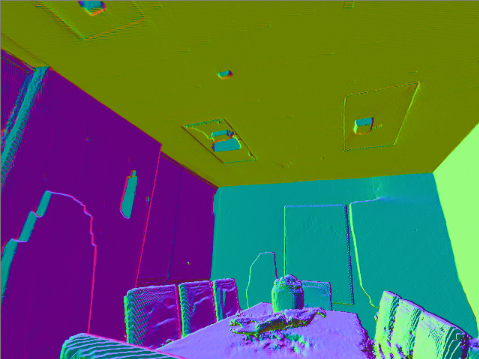}&
 \includegraphics[width=0.17\linewidth]{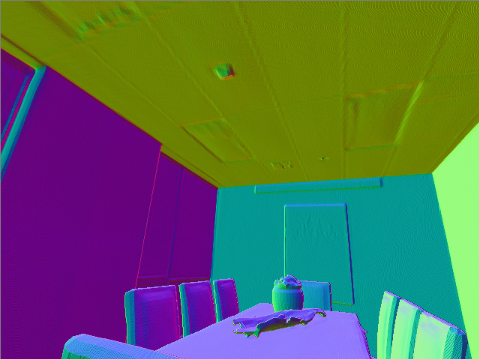} \\ \hline \hline
 
 \multirow{1}{*}[20pt]{\rotatebox{90}{RGB}} &
 \includegraphics[width=0.17\linewidth]{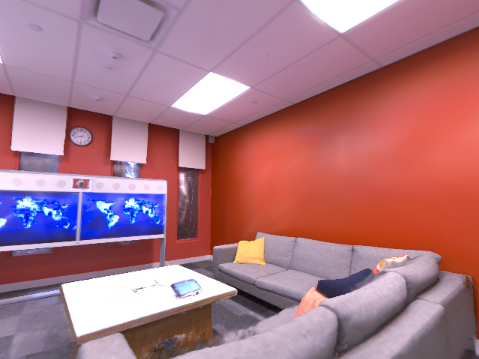}&
 \includegraphics[width=0.17\linewidth]{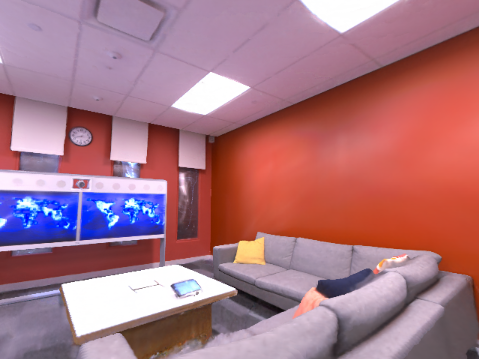}&
 \includegraphics[width=0.17\linewidth]{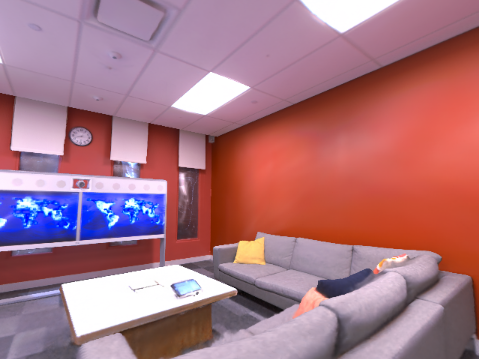}&
 \includegraphics[width=0.17\linewidth]{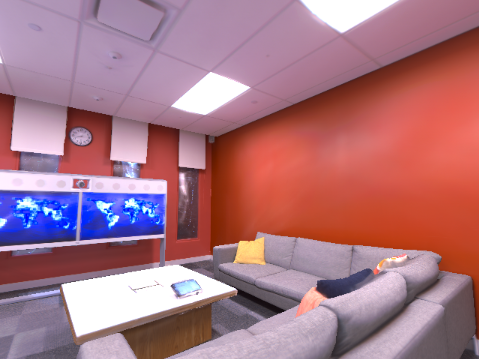} \\
 
 \multirow{1}{*}[30pt]{\rotatebox{90}{Normals}} &
 \includegraphics[width=0.17\linewidth]{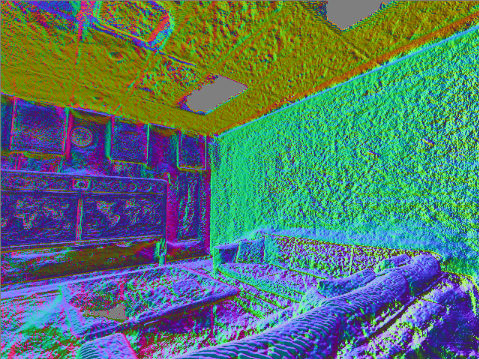}&
 \includegraphics[width=0.17\linewidth]{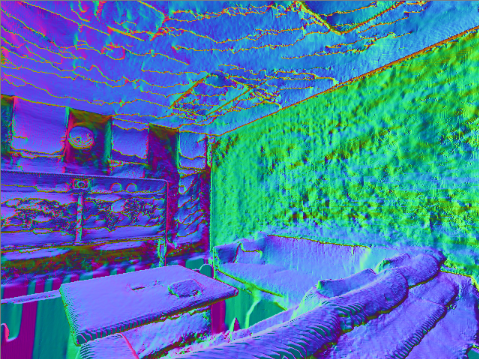}&
 \includegraphics[width=0.17\linewidth]{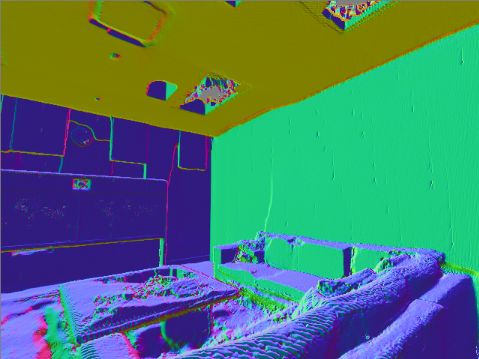}&
 \includegraphics[width=0.17\linewidth]{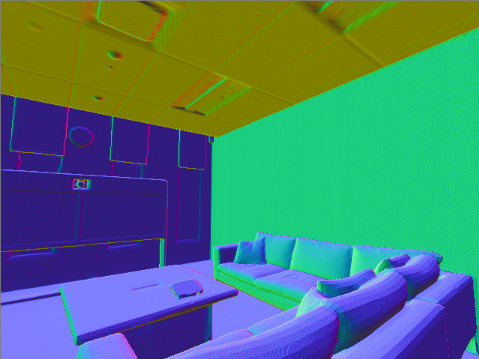} \\
 
\end{tabular}
	\caption{\textbf{Qualitative results on Replica.} Our method leverages many Manhattan objects and surfaces in the scene, which improves the implicit geometrical representation compared to the baseline. Unlike ManhattanDF, our method relies on many Manhattan cues other than the walls and floors.}
	\label{fig:supp:vis_replica2}
\end{figure*}

We provide more qualitative results related to experiments from
\ifx\FileIsMerged\undefined
Section 5 of the main paper.
\else
Section~\ref{sec:Experiments} of the main paper.
\fi
Figure~\ref{fig:supp:vis_hypersim} depicts visual results from the Hypersim-A dataset, Figure~\ref{fig:supp:scannet} depicts visual results from the ScanNet dataset, and Figure~\ref{fig:supp:vis_replica2}
depicts visual results from the Replica dataset.
We again see that our method leverages many Manhattan objects and surfaces in the scene, which improves the geometrical structure of 3D compared to the InstantNGP baseline. This is visible in surface normals and depth, obtained using volume rendering. Furthermore, unlike ManhattanDF, our method relies on many Manhattan cues other than the walls \& floors. For example, different component and parts of furniture and stairs respect the Manhattan assumption, which is successfully exploited by our method.



\clearpage
{\small
\bibliographystyle{ieee_fullname}
\bibliography{main}
}

\end{document}